\documentclass[letterpaper]{article} 
\usepackage{aaai24}  
\usepackage{times}  
\usepackage{helvet}  
\usepackage{courier}  
\usepackage[hyphens]{url}  
\usepackage{graphicx} 
\usepackage{natbib}  
\usepackage{caption} 
\usepackage{algorithm}
\usepackage{algorithmic}

\usepackage{appendix}
\usepackage{subfigure}
\usepackage{amsmath}
\usepackage{amssymb}
\usepackage{amsthm}

\def\E{{\mathbb{E}}}
\def\P{{\mathcal{P}}}
\def\A{{\mathcal{A}}}
\def\S{{\mathcal{S}}}

\newtheorem{theorem}{Theorem}
\newtheorem{proposition}[theorem]{Proposition}
\newtheorem{lemma}[theorem]{Lemma}

\usepackage{newfloat}
\usepackage{listings}
\DeclareCaptionStyle{ruled}{labelfont=normalfont,labelsep=colon,strut=off} 
\floatstyle{ruled}
\newfloat{listing}{tb}{lst}{}
\floatname{listing}{Listing}
\title{Relative Policy-Transition Optimization for Fast Policy Transfer}
\author{
    Jiawei Xu\textsuperscript{\rm 1,2}\equalcontrib,
    Cheng Zhou\textsuperscript{\rm 1},
    Yizheng Zhang\textsuperscript{\rm 1},
    Baoxiang Wang\textsuperscript{\rm 2},
    Lei Han\textsuperscript{\rm 1}\equalcontrib
}
\affiliations{
    \textsuperscript{\rm 1}Tencent Robotics X \\
    \textsuperscript{\rm 2}The Chinese University of Hong Kong, Shenzhen
    \\
    jiaweixu1@link.cuhk.edu.cn, \{mikechzhou,yizhenzhang\}@tencent.com \\
    bxiangwang@cuhk.edu.cn, lxhan@tencent.com

}

\usepackage{bibentry}

\begin{document}

\maketitle

\begin{abstract}

We consider the problem of policy transfer between two Markov Decision Processes (MDPs). We introduce a lemma based on existing theoretical results in reinforcement learning to measure the relativity gap between two arbitrary MDPs, that is the difference between any two cumulative expected returns defined on different policies and environment dynamics. Based on this lemma, we propose two new algorithms referred to as Relative Policy Optimization (RPO) and Relative Transition Optimization (RTO), which offer fast policy transfer and dynamics modelling, respectively. RPO transfers the policy evaluated in one environment to maximize the return in another, while RTO updates the parameterized dynamics model to reduce the gap between the dynamics of the two environments. Integrating the two algorithms results in the complete Relative Policy-Transition Optimization (RPTO) algorithm, in which the policy interacts with the two environments simultaneously, such that data collections from two environments, policy and transition updates are completed in one closed loop to form a principled learning framework for policy transfer. We demonstrate the effectiveness of RPTO on a set of MuJoCo continuous control tasks by creating policy transfer problems via variant dynamics.
\end{abstract}

\section{Introduction}
Deep reinforcement learning (RL) has demonstrated great successes in recent years, in solving a number of challenging problems like Atari~\citep{mnih2015human}, GO~\citep{silver2016mastering,silver2017mastering}, 
StarCraft II~\citep{astar,han2020tstarbot}, ViZDoom~\citep{li2023fittest} and legged robotics~\cite{han2023lifelike}.
These successes demonstrate that current deep RL methods are capable to explore and exploit sufficiently in huge observation and action spaces, as long as sufficient and effective data samples can be generated for training, such as the cases in games. For example, AlphaGo Zero~\citep{silver2017mastering} costs 3 days of training over 4.9 million self-play games, and OpenAI Five~\citep{OpenAI_dota} and AlphaStar~\citep{astar} spend months of training using thousands of GPUs/TPUs over billions of generated matches. 
However, for environments that prohibit infinite interactions, e.g., robotics, real life traffic control and autopilot, etc., applying general RL is difficult because generating data is extremely expensive and slow. Even if parallel data collection is possible, for example, by deploying multiple robots or vehicles running simultaneously, the scale of collected data is still far below that in virtual games. Worse still, exploration in these environments is considerably limited for safety reasons, which further reduces the effectiveness of the generated data. Due to the above challenges, similar significant advances like in solving virtual games have not been witnessed in these applications yet.

Generally, there are three tracks of approaches targeting to alleviate the aforementioned situation to promote the widespread application of RL. They are improving the data efficiency, transfer learning and simulator engineering.

To improve data efficiency, many recent efforts have been paid on investigating offline RL algorithms~\citep{rosete2023latent,zhan2022offline, lyu2022mildly,siegel2020keep,fujimoto2019off}. Compared to standard on-policy or off-policy RL, offline RL aims to effectively use previously collected experiences stored in a given dataset, like supervised learning, without online interactions with the environment. The stored experiences may not be generated by a fixed or known policy, so offline RL algorithms can leverage any previously collected data and learn a provably better policy than those who generated the experiences in the dataset. 
Although offline RL can effectively take advantage of finite data samples, solving a complex real-world task still requires a huge amount of high-quality offline experiences. Another way to increase data efficiency is to adopt model-based RL. Compared to model-free methods, model-based RL~\citep{bakker2001reinforcement,schaefer2007recurrent,kaiser2019model,janner2019trust,delecki2023model} learns a dynamics model that mimics the transitions in the true environment, and then the policy is free to interact with the learned dynamics. It has been proved that the true return can be improved by interacting with the learned dynamics model when the model error is bounded~\citep{janner2019trust}. However, learning an accurate dynamics model still requires sufficient transition by interacting with the true environment.

Transfer learning in RL~\citep{taylor2009transfer,zhu2023transfer} is practically useful to adapt a policy learned in a \emph{source} environment to solve another task in the \emph{target} environment. In the context of this paper, we consider the case \emph{where the policy is free to explore in the source environment, while the amount of collected data in the target environment should be as small as possible}. When the source environment is a simulated one while the target environment takes place in reality, the transfer problem is also known as the simulation to reality (sim2real) problem. The simplest transfer method is to train the policy in the source environment and then use the converged parameters as the warm start for a (or part of a) new policy in the target environment, so that the amount of interactions with the target is expected to be largely reduced, as long as the tasks and dynamics in the two environments are closely related.

An important concept in transfer learning is that instead of directly solving the target problem, a source task is considered in advance. Sharing this spirit, the last track of approaches tries to build a proxy simulator that is as close as possible to the target environment, and hence we refer to such methods as simulator engineering. For example, in robotics control problems, there are many mature toolboxes that can offer simulation engineering, including MuJoCo, PyBullet, Gazebo, etc. Model-based RL can also be viewed as a specific form of simulator engineering that the simulator is composed of a pure neural network, which is trained to approach the target environment as with lower a model error as possible, while this might require a large amount of dynamics data in the target environment as mentioned above. Actually, to achieve more efficient and accurate simulator engineering, one recent rising direction is to integrate differentiable programming and physical systems to build a trainable simulator, which follows the physical laws as in reality and also whose key factors, such as the mass, length or friction of some objects, are trainable like the parameters in neural networks. Representative examples include the DiffTaichi~\citep{hu2020difftaichi}, Brax~\citep{brax2021github} and Nimble~\citep{werling2021fast}.

Overall, the existing methods focus on either directly improving the data efficiency in the target environment or bridging/reducing the gap between a proxy environment and the target environment, and there lacks a principled way that can incorporate the learning in the two environments through a unified framework. 
In this paper, we inherit the spirit of transfer learning and consider two environments, where one is cheap to interact and another is the goal to solve, and the number of interactions in the target environment should be as small as possible. 
We believe that there exist some explicit connections between the expected returns in the two environments, given two different policies. 
Actually, previous theoretical approaches have analyzed the difference between two expected returns under different policies \citep{kakade2002,schulman2015trust}, or different dynamics \citep{luo2018algorithmic}, separately.

To provide an explicit value difference for the case where both the policy and dynamics vary in two Markov Decision Processes (MDPs), we introduce a lemma to combine the previous results in \citep{kakade2002,luo2018algorithmic}.
Let $\P(s'|s,a)$ and $\P'(s'|s,a)$ denote two dynamics transition functions in any two arbitrary MDPs sharing the same state and action spaces, where $(s,a,s')$ is the tuple of the state, action and next state. Let $\pi'(a|s)$ and $\pi(a|s)$ denote two arbitrary policies, and denote $J(\P, \pi)$ as the expected return given any $\P$ and $\pi$. 
Then, the lemma gives an explicit form for the value difference $J(\P',\pi)-J(\P,\pi')$, which is referred to as the \emph{relativity gap} between the two MDPs. 

Now, suppose $\P^{source}$ and $\P^{target}$ are the dynamics functions in the \emph{source} and \emph{target} MDPs respectively, and $J(\P^{source},\pi^{source})$ has been maximized by optimizing a parameterized $\pi^{source}$. 
Then, with fixed $\P^{source}$, $\P^{target}$ and $\pi^{source}$, maximizing the relativity gap over $\pi^{target}$ by constraining $\pi^{target}$ to be close to $\pi^{source}$ will also improve the return $J(\P^{target},\pi^{target})$; on the other hand, for trainable $\P^{source}$, minimizing the relativity gap by optimizing $\P^{source}$ given fixed policies $\pi^{source}=\pi^{target}$ will reduce the dynamics gap, similar to what is pursued by conventional model-based RL methods. 
Based on the above two principles and the value relativity lemma, we then propose two new algorithms referred to as Relative Policy Optimization (RPO) and Relative Transition Optimization (RTO), respectively. RPO transfers the policy evaluated in the source environment to maximize the return in the target environment, while RTO updates a dynamics model to reduce the value gap in the two environments. Then, applying RPO and RTO simultaneously offers a complete algorithm named Relative Policy-Transition Optimization (RPTO), which can transfer the policy from the source to the target smoothly. RPO, RTO and RPTO interact with the two environments simultaneously, so that data collections from two environments, policy and transition updates are completed in a closed loop to form a principled learning framework. 
In the experimental section, we show how to practically apply RPO, RTO and RPTO algorithms. We demonstrate the effectiveness of these methods in the MuJoCo continuous control tasks, by varying the physical variables of the objects to create policy transfer problems. In the last section, we discuss a few interesting directions which are worthy of future investigations.

\section{Preliminaries}
\textbf{Reinforcement Learning}.
A standard RL problem can be described by a tuple $\langle \mathcal{E}, \A, \S, \P, r, \gamma, \pi \rangle$, where $\mathcal{E}$ indicates the environment that is an MDP with dynamics transition probability $\P$; at each time step $t$, $s_t\in\S$ is the global state in the state space $\S$, and $a_t\in\A$ is the action executed by the agent at time step $t$ from the action space $\A$; the dynamics transition function $\P(s_{t+1}|s_t,a_t)$ is the probability of the state transition $(s_t,a_t)\xrightarrow{}s_{t+1}$; for the most general case, the reward $r(s_t,a_t,s_{t+1})$ can be written as a function of $s_t,a_t$ and $s_{t+1}$, while in many tasks it only relies on one or two of them, or it is even a constant in sparse rewards problem~\citep{xu2023efficient}. For notation simplicity, we usually write $r(s_t,a_t,s_{t+1})$ as $r_t$; $\gamma\in[0,1]$ is a discount factor and $\pi(a_t|s_t)$ denotes a stochastic policy. The following equations define some important quantities in reinforcement learning. 
The objective of RL is to maximize the expected discounted return
\begin{align}
J(\P,\pi)= & \mathbb{E}_{s_0,a_0,\cdots\sim\P,\pi}
\left[ \sum_{t=0}^{\infty}\gamma^t r_t \right], \\
\text{ where } s_0&\sim\P(s_0),\ 
a_t\sim\pi(a_t|s_t),\ s_{t+1}\sim\P(s_{t+1}|s_t,a_t). \nonumber
\end{align}
At time step $t$, the state-action value $Q^{\P,\pi}$, value function $V^{\P,\pi}$, and advantage $A^{\P,\pi}$ are defined as
\begin{align}
Q^{\P,\pi}(s_t,a_t)&=\E_{s_{t+1},a_{t+1},\cdots\sim\P,\pi}
\left[
\sum_{l=0}^{\infty}\gamma^l r_{t+l}
\right], \\
V^{\P,\pi}(s_t)&=\E_{a_{t},s_{t+1},\cdots\sim\P,\pi}
\left[
\sum_{l=0}^{\infty}\gamma^l r_{t+l}
\right],
\end{align}
and $A^{\P,\pi}(s,a)=Q^{\P,\pi}(s,a)-V^{\P,\pi}(s)$. 
In the above standard definitions, we explicitly show their dependence on both the dynamics $\P$ and policy $\pi$, since we will analyze these functions defined on variant dynamics and policies. This convention will be kept throughout the paper.

\textbf{The Policy Improvement Theorem}.
Given two arbitrary policies $\pi$ and $\pi'$, the policy improvement theorem~\citep{kakade2002,schulman2015trust} is the fact revealed by the following equation
\begin{equation}
\resizebox{.9\hsize}{!}{$
J(\P,\pi)=J(\P,\pi')+\E_{s_0,a_0,\cdots\sim\P,\pi}
\left[
\sum_{t=0}^\infty\gamma^t A^{\P,\pi'}(s_t,a_t)
\right].
$}
\label{eq:pit}
\end{equation}
Based on this theorem, some widely adopted RL algorithms such as TRPO~\citep{schulman2015trust} and PPO~\citep{schulman2017proximal} are developed.

\textbf{The Telescoping Lemma}.
The telescoping lemma \citep{luo2018algorithmic} reveals the value difference between any two dynamics functions $\P'$ and $\P$, given some fixed policy $\pi$. Using our notations, that is
\begin{equation}
\resizebox{.9\hsize}{!}{$
V^{\P',\pi}-V^{\P,\pi}=\frac{1}{1-\gamma}\big[
\E_{s'\sim\P'}V^{\P,\pi}(s')-\E_{s'\sim\P}V^{\P,\pi}(s')
\big].
$}
\label{eq:tele}
\end{equation}
Eq.~\eqref{eq:tele} was used to design the value difference bound and model-based RL algorithm in \citep{luo2018algorithmic}. In the following section, we will combine the policy improvement theorem in Eq.~\eqref{eq:pit} and the telescoping lemma in Eq.~\eqref{eq:tele} to obtain an explicit form of the value difference under the case for both different policies and dynamics functions.

\textbf{Soft Actor-Critic}.
SAC~\cite{haarnoja2018soft} is an off-policy RL algorithm. 
It aims to maximize the cumulative return as well as the policy entropy.
SAC uses the soft policy iteration to update the model parameters.
In the soft policy evaluation stage, it updates the soft Q-value with parameters $\mu$ by minimizing the soft Bellman residual
\begin{equation}
\begin{aligned}
    \mathcal{L}_Q(\mu) = \E_{(s_t, a_t) \sim \mathcal{D}} \left[( Q_{\mu}(s_t, a_t) - r_t - \gamma \bar{V}(s_{t+1}))^2\right], \\
    \text{with } \bar{V}(s_t) = \E_{a_t \sim \pi} \left[Q_{\bar{\mu}}(s_t, a_t) - \alpha \log \pi(a_t|s_t)\right],
\end{aligned}
\label{eq:soft_q}
\end{equation}
where $\mathcal{D}$ is the data replay buffer, $\bar{\mu}$ is the target Q parameters, and $\alpha$ is the temperature coefficient. 
In the soft policy improvement stage, the policy $\pi$ with parameters $\theta$ is updated by minimizing the objective
\begin{equation}
    \mathcal{L}_{\pi}(\theta) = \E_{s_t \sim \mathcal{D}} \bigg[ 
    \E_{a_t \sim \pi_{\theta}} \big[\alpha \log(\pi_{\theta}(a_t|s_t)) - Q_{\mu}(s_t, a_t)\big]
    \bigg].
\label{eq:ori_sac}
\end{equation}
The policy $\pi$ in SAC is a stochastic policy which is modelled as a Gaussian with mean and covariance given by the neural network for continuous action space. 
In this work, we develop our algorithm on the basis of SAC to handle continuous control tasks.

\section{The Value Relativity Lemma}
The following lemma integrates the policy improvement theorem with the telescoping lemma and measures the relative difference between any two expected returns under different policies and dynamics functions. 

\begin{lemma}
\label{theo:main}
Given two Markov Decision Processes (MDPs) denoted by $\mathcal{E}'$ and $\mathcal{E}$, who share the same state and action spaces $\S$, $\A$ and reward function $r$, their dynamics transition probabilities are defined as $\P'(s_{t+1}|s_t,a_t)$ and $\P(s_{t+1}|s_t,a_t)$ for any transition $(s_t, a_t)\xrightarrow{}s_{t+1}$ in $\mathcal{E}'$ and $\mathcal{E}$ respectively. Assume the initial state distributes identically in the two MDPs that $\P'(s_0)=\P(s_0)$. Let $J(\P,\pi)$ denote the expected return defined on dynamics $\P$ and policy $\pi$. Then, the relativity gap between any two expected returns under different dynamics and policies is defined as
\begin{equation}
\resizebox{.9\hsize}{!}{$
\underbrace{J(\P',\pi)-J(\P,\pi')}_{\text{relativity gap}}
= \\
\underbrace{J(\P',\pi)-J(\P,\pi)}_{\text{dynamics-induced gap}} +\underbrace{J(\P,\pi)-J(\P,\pi')}_{\text{policy-induced gap}}
$},
\label{eq:rg}
\end{equation}
such that the dynamics-induced gap can be derived from the telescoping lemma~\citep{luo2018algorithmic} in Eq.~\eqref{eq:tele} as
{\small
\begin{equation}
\begin{aligned}
J(&\P',\pi) - J(\P,\pi)= \\
&\E_{s_0,a_0,\cdots\sim \P',\pi}
\sum_{t=0}^\infty
\gamma^t
\big[
r_t
+\gamma V^{\P,\pi}(s_{t+1})-Q^{\P,\pi}(s_t,a_t)
\big],
\label{eq:did}
\end{aligned}    
\end{equation}
}and the policy-induced gap is revealed by the policy improvement theorem~\citep{kakade2002} in Eq.~(\ref{eq:pit}), rewritten by switching $\pi$ and $\pi'$ as
{\small
\begin{equation}
J(\P,\pi)-J(\P,\pi')=\E_{s_0,a_0,\cdots\sim\P,\pi}
\sum_{t=0}^\infty\gamma^t A^{\P,\pi'}(s_t,a_t).
\label{eq:pid}
\end{equation}
}
\end{lemma}
The proofs of Eqs.~\eqref{eq:did} and~\eqref{eq:pid} can follow from~\citep{luo2018algorithmic} and~\citep{kakade2002}, respectively, both using the telescoping expansion as the main technique.
In Appendix~\ref{sec:proof_relativity}, we also provide another complete proof by directly expanding the expected returns in Eq.~\eqref{eq:rg}.
Although these gaps in Lemma~\ref{theo:main} have been referred to as the policy improvement theorem and telescoping lemma in~\citep{luo2018algorithmic} and~\citep{kakade2002}, we use the terms value relativity gap, dynamics-induced gap and policy-induced gap, and refer to the above lemma as the \emph{Value Relativity Lemma} in this paper for uniformity.

Before proposing new algorithms based on the above lemma, we need to emphasize a few points implied in Eqs.~(\ref{eq:did}) and~(\ref{eq:pid}):
\begin{itemize}
	\item In Eq.~(\ref{eq:did}), the expectation is taken over the trajectory $s_0, a_0, \cdots$ sampled from the pair $(\P',\pi)$, while the value and state-action value functions in the expectation, i.e., $V^{\P,\pi}(s_{t+1})$ and $Q^{\P,\pi}(s_t,a_t)$, are defined on $(\P,\pi)$. This gives a practically useful hint that given a fixed policy $\pi$, the dynamics-induced gap can be calculated by measuring the value functions $V^{\P,\pi}(s_{t+1})$ and $Q^{\P,\pi}(s_t,a_t)$ in the dynamics $\P$ (imagining this is the source environment, where infinite data can be generated cheaply to accurately evaluate the value functions), while collecting (a few) data samples in $\P'$ (imagining this is the target environment) to estimate the expectation.
	\item In Eq.~(\ref{eq:did}), $\E_{s_{t+1}\sim\P'(\cdot|s_t,a_t)}[r_t+\gamma V^{\P,\pi}(s_{t+1})]\neq Q^{\P,\pi}(s_t,a_t)$, because the transition $(s_t,a_t)\xrightarrow{}s_{t+1}$ takes place in $\P'$ instead of $\P$, and hence Eq.~(\ref{eq:did}) is not zero whereas $\P'(s_{t+1}|s_t,a_t)\neq\P(s_{t+1}|s_t,a_t)$ happens for non-zero $r_t$ and $V^{\P,\pi}(s_{t+1})$ with high probability, especially for high-dimensional deep neural networks. 

\end{itemize}
In the following sections, we will introduce two new algorithms inspired by the Value Relativity Lemma, where one algorithm is for fast policy transfer from the source environment to the target environment, and another algorithm updates the parameterized dynamics in the source environment to be close to the dynamics in the target environment. Then, by combining the two algorithms, we obtain the complete algorithm to fast transfer a policy from the source environment to the target environment.

\section{Relative Policy Optimization (RPO)}
\label{sec:rpo}
As discussed previously, Eq.~(\ref{eq:did}) in the Value Relativity Lemma suggests a way of estimating the dynamics-induced value gap by evaluating $Q^{\P,\pi}(s_t,a_t)$ and $V^{\P,\pi}(s_{t+1})$ in $\P$ while sampling data in $\P'$, given $\pi$. Practically, it is of less interest to estimate the exact dynamics-induced gap. Instead, if we have trained a policy $\pi^*$ in $\P^{source}$ that maximizes $J(\P^{source},\pi)$, then we are interested in finding another $\hat{\pi}$ such that $\hat{\pi}=\arg\max_{\pi} [J(\P^{target},\pi)-J(\P^{source},\pi^*)]$, 
which improves $J(\P^{target},\pi)$. Normally, as long as $\P^{target}$ is not far from $\P^{source}$, finding $\hat{\pi}$ can use $\pi^*$ as a warm start. Motivated by this, we propose the following theorem to get a lower bound of the dynamics-induced value gap.

\begin{theorem}
\label{theo:2}
Define $D_{TV}^{max}(p,q)=\max_{x}D_{TV}(p(\cdot|x)||$ $q(\cdot|x))$ as the total variation divergence between two distributions $p(\cdot|x)$ and $q(\cdot|x)$, 
where $D_{TV}(p(\cdot|x)||q(\cdot|x))=\frac{1}{2}\sum_{y}|p(y|x)-q(y|x)|$. 
Define $\epsilon=\max_{s,a}|A^{\P,\pi}(s,a)|$, where $A^{\P,\pi}(s,a)=Q^{\P,\pi}(s,a)-V^{\P,\pi}(s)$ is the advantage. 
Let $\delta_1=D_{TV}^{max}(\P'(\cdot|s,a),\P(\cdot|s,a))$
for any $\P'$, $\P$, and $(s,a)\in\S\times\A$, and let 
$\delta_2=D_{TV}^{max}(\pi'(\cdot|s),\pi(\cdot|s))$
for any $s\in\S$, and policies $\pi'$ and $\pi$. Let $r_{max}=\max_{s,a,s'}r(s,a,s')$ be the max reward for all $(s,a,s')$.\footnote{The above terms and assumptions have been widely used in RL in the literature such as \citep{kakade2002,schulman2015trust,luo2018algorithmic,janner2019trust}.}
Let $\Delta^{\P',\P}(\pi)=J(\P',\pi)-J(\P,\pi)$ be a function of $\P'$, $\P$ and $\pi$. 
Now, we import a new policy $\pi'$ and define the following function
{\small
\begin{align}
\sum_{t=0}^{\infty}\gamma^t\ 
\E_{s_t\sim \P',\pi'}
\sum_{a_t}\pi(a_t|s_t)
\sum_{s_{t+1}}\P'(s_{t+1}|s_t,a_t)  \cdot \nonumber \\
\left[
r(s_t,a_t,s_{t+1})+\gamma V^{\P, \pi'}(s_{t+1}) - Q^{\P, {\pi'}}(s_t,a_t)
\right],
\end{align}
}as an approximation of $\Delta^{\P',\P}(\pi)$ by both sampling $s_0,a_0,$ $\cdots,s_t$ and evaluating $V^{\P,\pi'}$ and $Q^{\P,\pi'}$ using $\pi'$. Then, we have the following lower bound
{\small
\begin{equation}
\begin{aligned}
\Delta^{\P',\P}(\pi)&\geq L_{\pi'}(\pi)-C_1, \\
\text{where }
C_1&=\frac{4\gamma r_{max}\delta_1}{(1-\gamma)^2}
\min\left(
\frac{\delta_2(\gamma^2+2)}{1-\gamma}, 1+\frac{\delta_2}{1-\gamma}
\right).
\label{eq:lower}
\end{aligned}
\end{equation}
}
\end{theorem}
The proof is provided in Appendix~\ref{sec:proof_theorem1}. In Theorem~\ref{theo:2}, we import $L_{\pi'}(\pi)$ defined on two policies $\pi$ and $\pi'$, because in practice we will need an algorithm iterating over the current policy parameter and its old parameter since last update. This technique will be adopted multiple times in the following context. Based on Theorem~\ref{theo:2}, we can further obtain the following lower bound of the entire relativity gap.

\begin{proposition}
\label{prop:1}
The entire relativity gap in Eq.~(\ref{eq:rg}) has the following lower bound
{\small
\begin{align}
J(\P',\pi)&-J(\P,\pi')\geq 
\frac{1}{1-\gamma}\E_{s\sim d^{\P',\pi'} ,a,s'\sim\P',\pi'}\frac{\pi(a|s)}{\pi'(a|s)}\cdot \nonumber
\\
&[r(s,a,s')+\gamma V^{\P,\pi'}(s')-V^{\P,\pi'}(s)]
-C_2,
\end{align}
}where $s'$ is the next state that $(s,a)\xrightarrow{}s'$, and
\begin{equation}
\begin{aligned}
\resizebox{0.9\hsize}{!}{$
C_2=\frac{2\gamma\epsilon(\delta_1+2\delta_2^2)}{(1-\gamma)^2}
+
\frac{4\gamma r_{max}}{(1-\gamma)^2}
\cdot\min
\left(
\frac{\delta_2(\gamma^2+2)}{1-\gamma}, 1+\frac{\delta_2}{1-\gamma}
\right),
$}
\end{aligned}
\end{equation}
is a constant relying on the dynamics discrepancy $\delta_1$ and policy discrepancy $\delta_2$.
\end{proposition}
The proof is provided in Appendix~\ref{sec:proof_prop1}.
By taking $\pi=\pi_{\theta}$ and $\pi'=\pi_{\theta_{old}}$ for some policy parameters $\theta$ and its old version $\theta_{old}$ since last update, Proposition~\ref{prop:1} suggests maximizing the following objective
{\small
\begin{align}
\E_{
\atop{
s\sim d^{\P',\pi_{\theta_{old}}},
\atop a,s'\sim\P',\pi_{\theta_{old}}}
}
\frac{\pi_{\theta}(a|s)}{\pi_{\theta_{old}}(a|s)}\cdot
[r(s,a,s')+\gamma V^{\P,\pi_{\theta_{old}}}(s')],
\label{eq:rpo}
\end{align}
}by noting that $V^{\P,\pi'}(s)$ serves as a baseline and does not affect the policy gradient.
Eq.~\eqref{eq:rpo} is very similar to Eq. (15) in~\cite{schulman2015trust}, but
recall that $r(s,a,s')+\gamma V^{\P,\pi_{\theta_{old}}}(s')
\neq Q^{\P,\pi_{\theta_{old}}}(s,a)$,
because the transition $(s,a)\xrightarrow{}s'$ takes place in $\P'$ instead of $\P$. 
Empirically, we consider a local approximation for Eq.~\eqref{eq:rpo} as
{\small
\begin{align}
&\E_{
\atop{s\sim d^{\P',\pi_{\theta_{old}}},
\atop a\sim\pi_{\theta_{old}},
s'\sim\P}
}
\frac{\pi_{\theta}(a|s)}{\pi_{\theta_{old}}(a|s)}\cdot
[r(s,a,s')+\gamma V^{\P,\pi_{\theta_{old}}}(s')]
\nonumber
\\
=\ &\E_{
\atop{s\sim d^{\P',\pi_{\theta_{old}}},
\atop a\sim\pi_{\theta},
s'\sim\P}
}
Q^{\P,\pi_{\theta_{old}}}(s,a),
\label{eq:sac}
\end{align}
}by approximating only one step transition $(s,a)\xrightarrow{}s'$ in $\P$. Note that Eq.~\eqref{eq:sac} still differs from standard Q-value maximization that $s$ is sampled from $\P'$ while the Q-value is computed in $\P$. 
Now, to optimize the policy through maximizing Eq.~\eqref{eq:sac}, we follow Eq.~\eqref{eq:ori_sac} in SAC by sampling $s$ from $\P'$ while estimating the Q-value in $\P$.
We refer to this procedure as \emph{Relative Policy Optimization} (RPO).

Now, we still need to look into the tightness of the bound in Proposition~\ref{prop:1}. In this section, we have considered fixed and diverse $\P$ and $\P'$, and hence $\delta_1$ is always a positive constant. Then, $C_2$ in Proposition~\ref{prop:1} will never approach zero by noting that $C_2>\frac{2\gamma\epsilon}{(1-\gamma)^2}\delta_1>0$ even if $\delta_2$ is sufficiently small, because the dynamics discrepancy prevents this. With the existence of $\delta_1>0$, $J(\P',\pi_{\theta})$ is improved over $J(\P,\pi_{\theta_{old}})$ (a constant at the current step) only when we can improve the objective in Eq.~\eqref{eq:rpo} by at least $(1-\gamma)C_2$.
Worse still, starting from a well trained policy $\pi_{\theta^*}$ in $\P$, as RPO updates $\theta$, $\pi_{\theta_{old}}$ will be gradually far from $\pi_{\theta^*}$, and therefore $J(\P,\pi_{\theta_{old}})$ might decrease. If this happens, maximizing the value gap $J(\P',\pi_{\theta})-J(\P,\pi_{\theta_{old}})$ will not guarantee the increase of $J(\P',\pi_{\theta})$. 
The ideal case is that RPO still keeps $\pi_{\theta}$ perform well in $\mathcal{E}$. 
As we can imagine, if this happens, $\pi_{\theta}$ will be updated towards a robust policy that performs well in both $\mathcal{E}'$ and $\mathcal{E}$. 
Indeed, as we will show in our experiments, as long as $\P'$ is not too far away from $\P$, optimizing RPO is able to obtain such a robust policy; however, once $\P'$ differs from $\P$ too much, RPO will fail to transfer the policy to the target environment. This can be remarked as a disadvantage of the RPO algorithm which requires a relatively small dynamics gap, i.e., $\delta_1$.
Overall, to guarantee the success of policy transfer, we need further to eliminate the dynamics discrepancy. 
This is possible when $\P$ (the dynamics in the source environment) is trainable, as considered in physical dynamics modeling~\citep{hu2020difftaichi,brax2021github,werling2021fast} and model-based RL methods~\citep{janner2019trust}. In the next section, we will consider a trainable $\P$.

\section{Relative Transition Optimization (RTO)}
\label{sec:rto}
In this section, given fixed $\pi$ and $\P'$, we consider a trainable $\P$. Suppose $\P_{\phi}(s'|s,a)$ is parameterized by $\phi$ for any transition $(s,a)\xrightarrow{}s'$. In the following theorem, we will import three dynamics quantities $\P'$, $\P_{\phi}$ and $\P_{\phi'}$, where $\P'$ can be treated as the dynamics in the target environment, and $\P_{\phi}$ and $\P_{\phi'}$ are two variant dynamics functions parameterized by $\phi$ and $\phi'$, respectively.

\begin{theorem}
\label{theo:3}
With the definitions in Theorem~\ref{theo:2}, introduce the following function
{\small
\begin{align}
L_{\phi'}(\phi)
=\sum_{t=0}^{\infty}&\gamma^t\ 
\E_{\tau \sim \P',\pi}
\bigg[
\E_{s_{t+1}\sim\P'}\big[
r_t+\gamma V^{\P_{\phi'},\pi}(s_{t+1})
\big]-
\nonumber \\
&\E_{s_{t+1}\sim\P_{\phi}}
\big[
r_t+\gamma V^{\P_{\phi'},\pi}(s_{t+1})
\big]
\bigg],
\end{align}
}as an approximation of $\Delta^{\P',\P_{\phi}}(\pi)$ by evaluating the value $V$ using $\P_{\phi'}$ instead of $\P_{\phi}$. Then, we have
{\small
\begin{equation}
\begin{aligned}
|\Delta^{\P',\P_{\phi}}(\pi)|
\leq
|L_{\phi'}(\phi)|+
\frac{4\gamma\delta_1r_{max}}{(1-\gamma)^2}
\min\left(\frac{\delta_2(\gamma^2+1)}{1-\gamma}, 1\right).
\end{aligned}
\end{equation}
}
\end{theorem}

In order to reduce the dynamics-induced gap by updating $\phi$, we have to minimize $|\Delta^{\P',\P_{\phi}}(\pi)|$. 
Based on Theorem~\ref{theo:3}, we can alternatively minimize $|L_{\phi'}(\phi)|$. Empirically, if we consider $\P_{\phi}$ as a probability function, then by noting that
{\small
\begin{align}
L_{\phi'}(\phi)=
\sum_{t=0}^{\infty}
\gamma^t\ 
\E_{\tau \sim \P',\pi}
\sum_{s'}
\big(
&\P'(s'|s,a) - \P_{\phi}(s'|s,a)
\big) \cdot  \nonumber \\
\big(
&r+\gamma V^{\P_{\phi'},\pi}(s')
\big),
\end{align}
}and taking $\phi'$ as the old version of $\phi$ since the last update, we consider a square loss on minimizing $|L_{\phi'}(\phi)|$ as
{\small
\begin{align}
\text{minimize}_{\phi}\ 
\E_{\tau \sim \P',\pi}
\sum_{s'} 
\big(\P'(s'|s,a)-&\P_{\phi}(s'|s,a)\big)^2 \cdot \nonumber \\
\big(
&r+\gamma V^{\P_{\phi'},\pi}(s')
\big)^2.
\label{eq:rto_p1}
\end{align}
}That is, we sample $(s,a)$ in $\mathcal{E}'$ using $\pi$ and optimize Eq.~(\ref{eq:rto_p1}). 
We refer to the above optimization as \emph{Relative Transition Optimization} (RTO).
It is easy to see that RTO implies the standard model-based RL methods, which directly train $\phi$ using supervised learning, i.e.,
{\small
\begin{equation}
\begin{aligned}
\text{minimize}_{\phi} \ \E_{\tau \sim \P',\pi}\sum_{s'}
\left(\P'(s'|s,a)-\P_{\phi}(s'|s,a)\right)^2.
\end{aligned}
\end{equation}
}Indeed, the objective of RTO in Eq.~(\ref{eq:rto_p1}) can be viewed as a weighted form of supervised learning, with the weight $(r+\gamma V^{\P_{\phi'},\pi}(s'))^2$ showing that transitions with larger values evaluated by $\P_{\phi'}$ and $\pi$ should be optimized more aggressively. This is reasonable from the perspective of fitting state values, instead of purely fitting the transition dynamics. Therefore, RTO provides a theoretical explanation for the standard model-based RL methods by reducing the dynamics-induced value gap, and absorbs standard model-based RL as a special case in RTO.

In this work, we learn the soft value function by following SAC~\cite{haarnoja2018soft}.
Then, we give the final objective for minimizing the dynamics-induced gap as
{\small
\begin{equation}
\begin{aligned}
&\E_{\tau \sim \P',\pi}
\sum_{s'} 
\big( \P'(s'|s,a)- \P_{\phi}(s'|s,a)\big)^2 w(r, s') 
\\
&\text{with }w(r, s') = \big(
r+\gamma (Q^{\P_{\phi'},\pi}(s', \pi(s')) - \log \pi(s') )
\big)^2,
\end{aligned}    
\label{eq:rto_p2}
\end{equation}
}where the Q-value will be optimized to minimize the soft Bellman residual.
In our implementation, we normalize the dynamics weight $w$ in the replay buffer $\mathcal{D}_{source}$ to guarantee numerical stability by
    $\hat{w} = \frac{w - w_{min} }{w_{max} - w_{min}} + \epsilon,$
where $w_{min}$ and $w_{max}$ are the smallest and largest weights in the buffer $\mathcal{D}_{source}$, 
and $\epsilon$ controls the minimum weight for updating.

In the experiments, we try both probabilistic and deterministic dynamics models. For the latter case, we adopt a deterministic physical dynamics model, similar to recently proposed differentiable  simulators~\citep{hu2020difftaichi,brax2021github,werling2021fast}, which take advantage of the inherent physical processes. Such a model is much more efficient than a pure neural network based dynamics model, because only a few scalars 
such as mass, length, gravity, etc., 
are trainable parameters in the physical systems. However, differentiable  simulator requires specific engineering on the considered environments and is not general for broad usage.

\section{Relative Policy-Transition Optimization (RPTO) Algorithm}
\label{sec:rpto}
Combining RPO and RTO, we finally obtain the complete Relative Policy-Transition Optimization (RPTO) algorithm.
We provide the overall algorithm of RPTO in Algorithm~\ref{alg:rpto}.

\begin{algorithm}[tb]
   \caption{Relative Policy-Transition Optimization.}
   \label{alg:rpto}
\begin{algorithmic}
   \STATE {\bfseries Input:} The source and target environments $\mathcal{E}^{source}$ and $\mathcal{E}^{target}$, and their dynamics $\P_{\phi_0}^{source}$ and $\P^{target}$, where $\phi_0$ can accurately describe the initial source dynamics; a well-trained policy $\pi_{\theta_0}$ in $\mathcal{E}_{\phi_0}^{source}$;
   \STATE \textbf{1.} Create two empty replay buffers $\mathcal{D}_{source}$ and $\mathcal{D}_{target}$;
   \STATE \textbf{2.} Initialize $\theta=\theta_0$ and $\phi=\phi_0$;
   \REPEAT
   \STATE \textbf{3.} Using $\pi_{\theta}$ to interact with $\mathcal{E}_{\phi}^{source}$ and push the generated trajectories into $\mathcal{D}_{source}$;
   \STATE \textbf{4.} Using $\pi_{\theta}$ to interact with $\mathcal{E}^{target}$ and push the generated trajectories into $\mathcal{D}_{target}$;
   \STATE \textbf{5.} Sample a mini-batch $\{(s,a,s')\}_{source}\sim\mathcal{D}_{source}$, then update $Q^{\P_{\phi}^{source},\pi_{\theta}}$ by minimizing the soft Bellman residual;
   \STATE \textbf{6.} Sample a mini-batch $\{(s,a,s')\}_{target}\sim\mathcal{D}_{target}$, then apply RPO and RTO to update $\pi_{\theta}$ and $\P_{\phi}^{source}$ respectively;
   \UNTIL{Some convergence criteria is satisfied.}
\end{algorithmic}
\end{algorithm}

As we observe in Algorithm~\ref{alg:rpto}, in the main loop of RPTO, the policy $\pi_{\theta}$ interacts with the two environments simultaneously and pushes the data into two buffers separately. In step 5, we sample a mini-batch from $\mathcal{D}_{source}$ to update the soft Q-value in Eq.~(\ref{eq:soft_q}). In step 6, we sample a mini-batch from $\mathcal{D}_{target}$ to update the policy parameter $\theta$ according to Eq.~(\ref{eq:rpo}) and Eq.~\eqref{eq:ori_sac} in RPO, and also update the dynamics $\phi$ according to Eq.~(\ref{eq:rto_p2}) from RTO. Therefore, RPTO combines data collection from two environments, RPO and RTO in a closed loop, and this offers a principled learning framework for policy transfer. Indeed, steps 5 and 6 can be parallelized, as long as the Q-value function $Q^{\P_{\phi}^{source},\pi_{\theta}}$ (suppose it is parameterized by $\mu$) does not share parameters with $\pi_{\theta}$, i.e., $\mu$ and $\theta$ are independent. Also, $\mu$ can be updated more frequently because it only requires data generated from the source environment, and the more accurate the Q-value function in the source environment is, the more accurate the relative policy gradient in step 6 will be estimated. 
For the dynamics parameter $\phi$, the case turns out to be different, and it is often not a good choice to update $\phi$ as fast as we can. To understand this, we need to explain how RPTO transfers the policy to the target environment in advance. 

As we can imagine, as $\P_{\phi}^{source}$ approaches $\P^{target}$ gradually by updating $\phi$, the policy $\pi_{\theta}$ is able to interact with a sequence of smoothly varying `source' environments. During this training period, the policy observes much more diverse transitions that lie between the initial source environment (with dynamics $\P_{\phi_0}^{source}$) and the target environment. These diverse transitions are very helpful to encourage the agent to explore a robust policy. On the other hand, since the policy is initialized with a well-trained $\theta_0$ in $\P_{\phi_0}^{source}$, smoothly varying $\P_{\phi_0}^{source}$ to reach $\P^{target}$ generates a sequence of environments that naturally provide a curriculum learning scheme, and this is very similar to what is considered in the recently emerged Environment Design~\citep{dennis2020emergent}. Now, we could answer the question in the last paragraph that why $\phi$ should not be updated aggressively: a slowly and smoothly varying $\phi$ provides more chances for the agent to see diverse transitions. Actually, if $\P_{\phi}^{source}$ approaches $\P_{target}$ too fast, RPTO will be similar to directly training SAC in the target environment with a warm start. 
In our implementation, we apply fewer dynamics updates than the original MBPO~\cite{janner2019trust}.

\section{Experiments}

In this section, we perform comprehensive experiments across diverse control tasks to assess the efficacy of our RPTO. Additionally, we analyze how RPT, RTO, and RPTO individually contribute to learning.

\begin{figure}[t]
\centering
\subfigure[Source]{\includegraphics[width=0.45\linewidth]{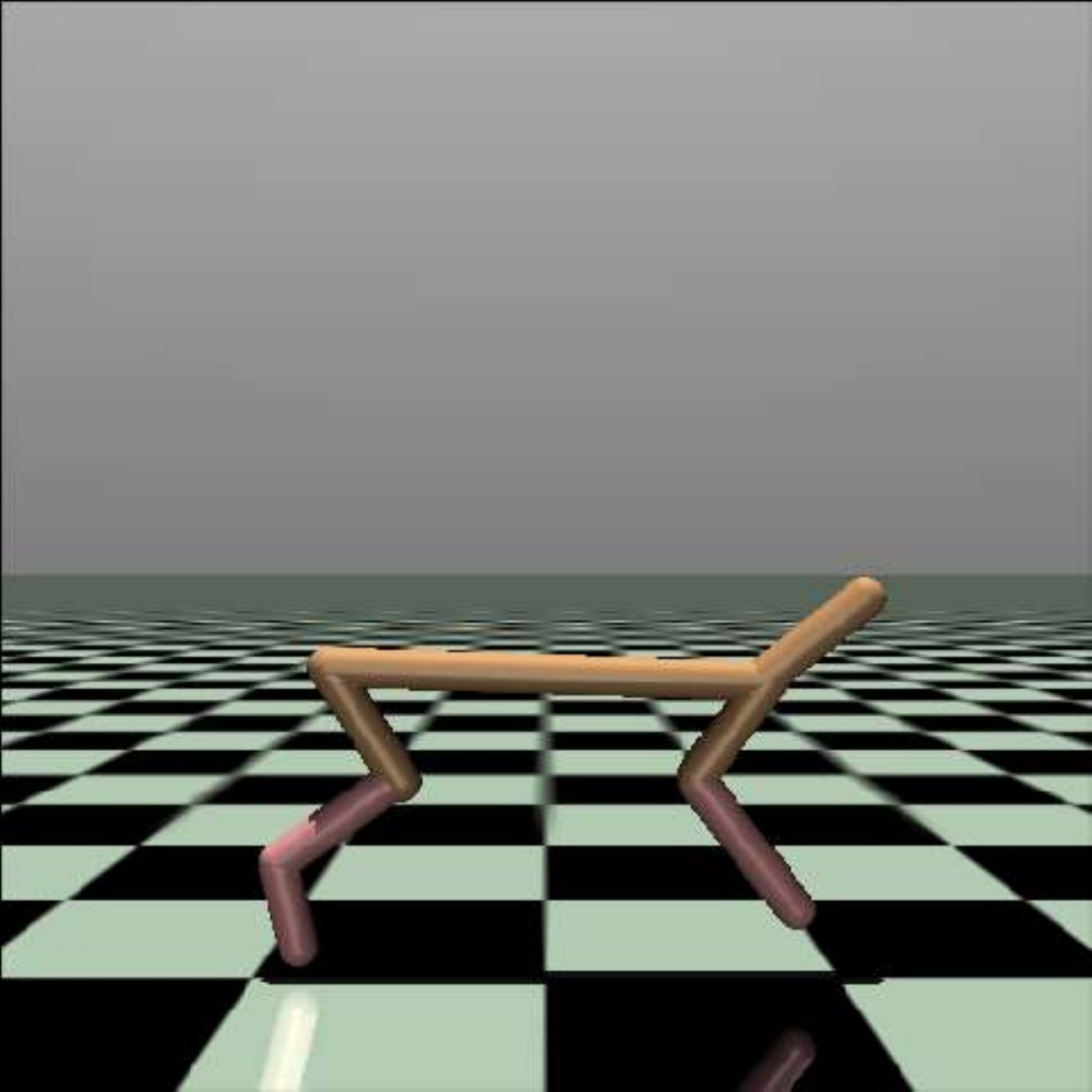}}
\subfigure[Target]{\includegraphics[width=0.45\linewidth]{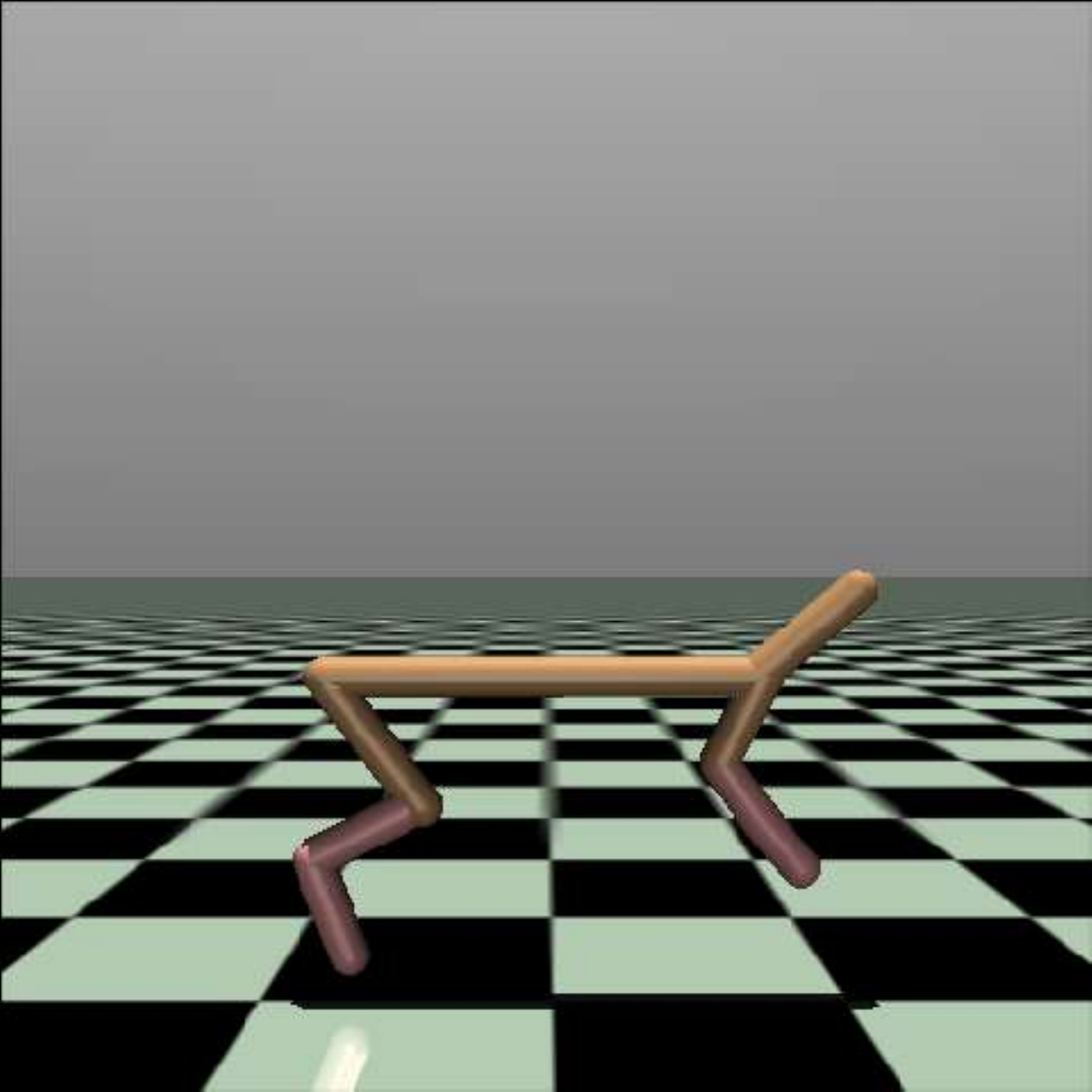}}
\caption{An illustration of the source and target environments on HalfCheetah.
}
\label{fig:half}
\end{figure}

\textbf{Environment Settings.}
We experiment on a set of MuJoCo continuous control tasks with the standard neural network (NN) based probabilistic dynamics model, including Ant, Hopper, HalfCheetah, Walker2d and Swimmer.
We first modify the default observation space and action space in some tasks as the source environments,
while for each task, we arbitrarily modify some of its physical factors to create the corresponding target environments. 
For MuJoCo tasks, we randomly add a $\pm20\%$ variation to the length of all joints of the robot and the friction of the environment. 
We use the default setting in OpenAI gym as our source environment.
Figure~\ref{fig:half} shows an illustration of the source and target environments.
All joints in the target environment have different lengths than in the source environment.
Please see Appendix~\ref{sec:G} for detailed settings on environments.

\textbf{Baselines}.
We first pre-train a converged policy and dynamics using the pure model-based algorithm in the source environment, and the pre-trained policy and dynamics will be used as a warm start when transferring to the target environment. For the policy transfer stage, SAC~\cite{haarnoja2018soft} and TRPO~\cite{schulman2015trust} with the pre-trained policy as initialization, denoted as SAC-warm and TRPO-warm respectively, will be used as the baseline methods. 
We also involve the state-of-the-art model-based methods MBPO~\citep{janner2019trust}, SLBO~\cite{luo2018algorithmic}, PTP~\cite{wu2022plan} and PDML~\cite{wang2023live}, as the baseline methods. Both of them use their pre-trained model from the source environment as initialization to achieve fair comparison in transfer tasks.
All the compared algorithms are open-sourced, thus we can use their original implementations to avoid biased evaluation.

\begin{figure*}[ht]
\centering
\includegraphics[width=1.0\linewidth]{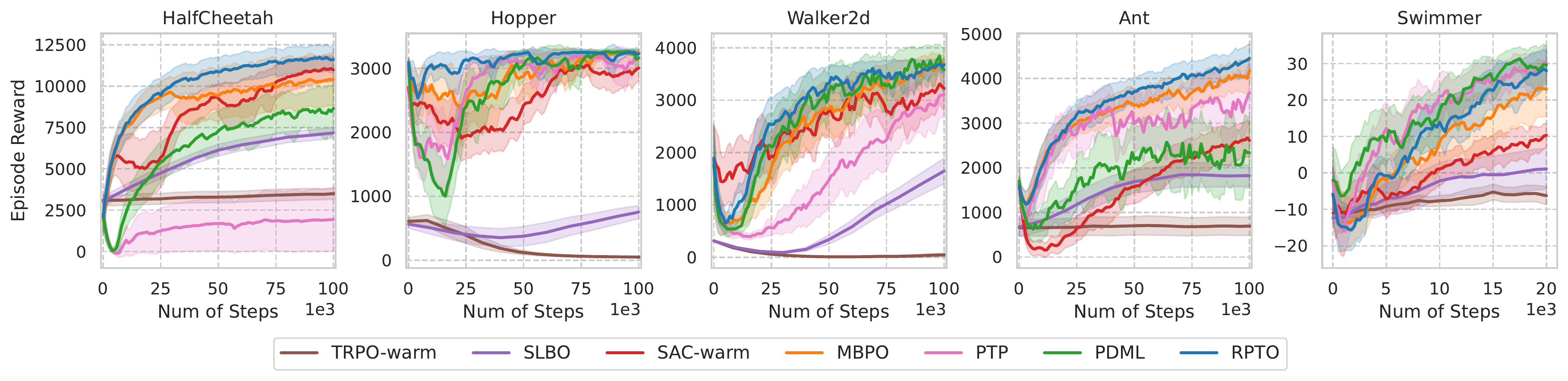}
\caption{Overall performance on MuJoCo tasks.}
\label{fig:mujoco}
\end{figure*}

\textbf{Practical Implementation}.
We build the neural dynamics model by following MBPO, and use an ensemble network to learn the dynamics.
We set the ensemble size to 7 and each ensemble network has 4 fully connected layers with 400 units.
Each head of dynamics model is a probabilistic neural network which outputs Gaussian distribution with diagonal covariance: $p_{\phi}^i(s_{t+1}, r_t|, s_t, a_t) = \mathcal{N}\big(\mu_{\phi}^i(s_t, a_t), \sum_{\phi}^i(s_t,a_t)\big)$. We set the model horizon to 1 and the replay ratio of dynamics to 1 for all environments.
Other implementation details are provided in Appendix~\ref{sec:G}. 
For every environment, we create 4 different target environments, and we repeat 4 times for each target environment.
Therefore, we totally repeat 16 times for each environment to compute the mean curve with $95\%$ confidence interval.

\textbf{Benchmark Results}.
We show the results of all methods on the MuJoCo tasks in Figure~\ref{fig:mujoco}. 
In these environments, we first pre-train the policy and the neural network (NN) dynamics model with MBPO and SLBO. Then, we use them as warm start in the RPTO, MBPO and SLBO methods. 
Moreover, we use the pre-trained policy in MBPO and SLBO as warm start for SAC-warm and TRPO-warm, respectively.
As we can see, the model-based method MBPO can serve as a strong baseline for policy transfer. 
However, in all the tasks, the RPTO methods demonstrate superiority over the other baselines in terms of both fast policy transfer and even better asymptotic convergence, because RPTO sees much more diverse dynamics that promote exploration, as explained at the end of the RPTO description section.

\textbf{Didactic Example}.
We then use CartPole as an illustrative environment to show how the proposed algorithms are practically useful. 
In this experiment, we evaluate the RPO, RTO and RPTO with deterministic physical dynamics model, because the physical systems are explicitly coded in CartPole, allowing us to conveniently build physical dynamics model with only a few trainable factors.
In this didactic example, only the pole length is treated as the trainable parameter, i.e., $\phi$ in CartPole only contains one free parameter. 
Similarly, we also first pre-train a converged policy in the source environment, but we don't need to pre-train the dynamics model this time.
We use deterministic physical dynamics to minimize the dynamics gap between the source and the target environment.

\begin{figure}[t]
\centering
\includegraphics[width=1.0\columnwidth]{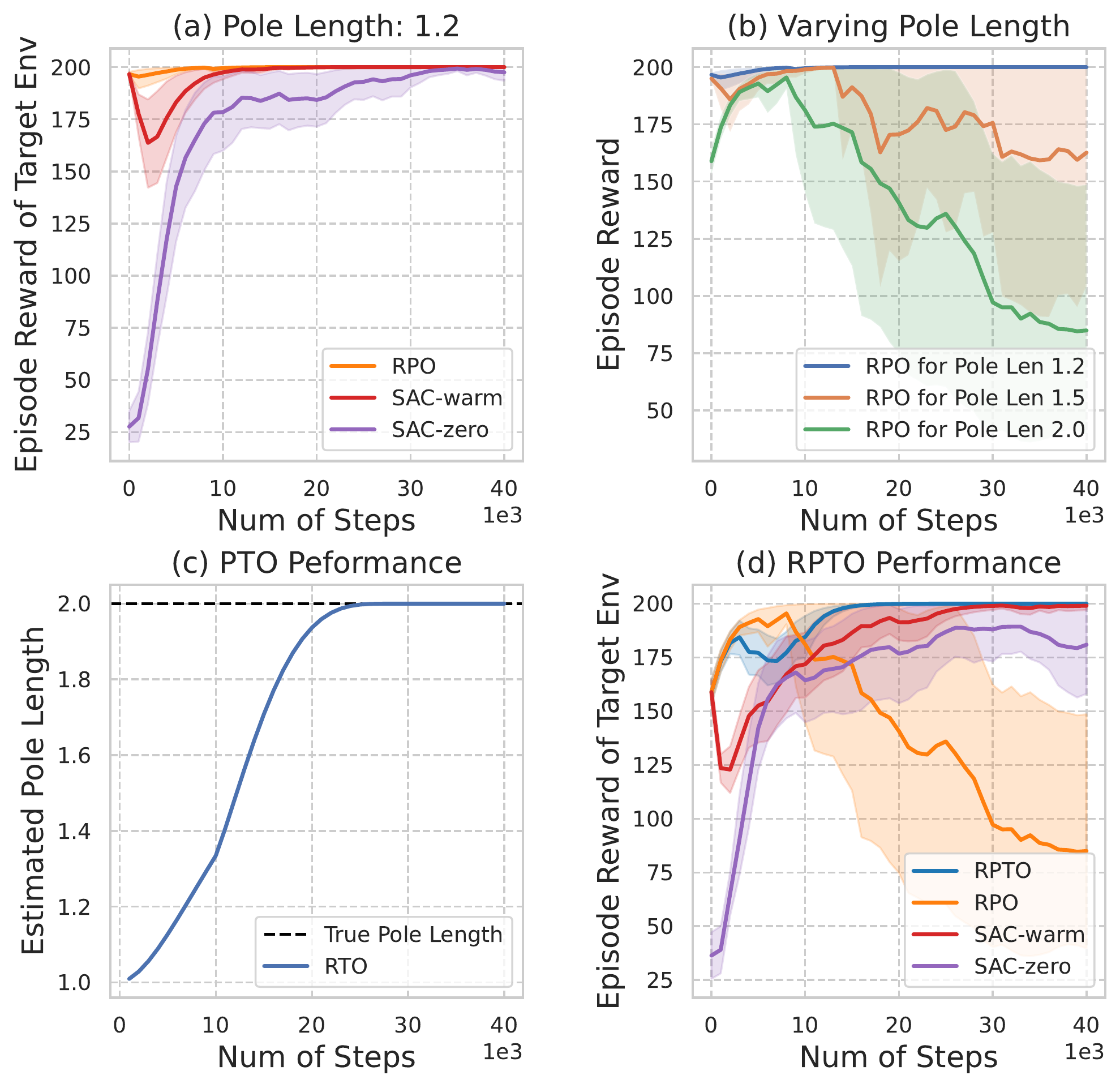}
\caption{Illustrative experiments in CartPole. The length of pole in source environment is 1.0.
}
\label{fig:cartpole}
\end{figure}

Figure~\ref{fig:cartpole} reports the evaluation results of our RPO, RTO and RPTO. Figure~\ref{fig:cartpole}(a) demonstrates that RPO (without RTO) is sufficient to successfully transfer the policy to the target environment with a pole length of 1.2, and RPO transfers the policy faster than SAC-warm. In Figure~\ref{fig:cartpole}(b), by varying the pole length to test more target environments, we find that when the pole length difference becomes larger, RPO fails to obtain a stable policy in the target environment. 
These results are consistent with our analysis at the RPO section.
Figure~\ref{fig:cartpole}(c) evaluates the performance of RTO. As we can observe, RTO can optimize $\phi$ to converge to the true pole length in the target environment. 
Finally, Figure~\ref{fig:cartpole}(d) reports the RPTO's performance on the tagret environment with the pole length of 2.0.
As we can observe, RPTO transfers the policy much faster than all the other methods.

\section{Future Work}
Our work is related to model-based RL, environment design and sim2real problems. 
There are some future directions worth investigating. 
For example, as we have discussed before, controlling the update step size or frequency of RTO can provide better curriculum learning or environment design (although in this paper we simply fix the learning rate of RTO). Connecting this with meta-RL and current environment design algorithms is a promising future direction. Moreover, the theory and algorithms in this paper are orthogonal to other techniques commonly used in sim2real tasks, such as domain adaptation, domain randomization, augmented observation, etc., and all of these methods can be integrated together into sim2real applications. 

\section{Acknowledgments}
Baoxiang Wang is partially supported by National Natural Science Foundation of China (62106213, 72150002, 72394361) and Shenzhen Science and Technology Program (RCBS20210609104356063, JCYJ20210324120011032).

\bibliography{ref}

\begin{thebibliography}{32}
\providecommand{\natexlab}[1]{#1}

\bibitem[{Bakker(2001)}]{bakker2001reinforcement}
Bakker, B. 2001.
\newblock Reinforcement learning with long short-term memory.
\newblock \emph{Advances in Neural Information Processing Systems (NeurIPS)},
  14.

\bibitem[{Berner et~al.(2019)Berner, Brockman, Chan, Cheung, Debiak, Dennison,
  Farhi, Fischer, Hashme, Hesse et~al.}]{OpenAI_dota}
Berner, C.; Brockman, G.; Chan, B.; Cheung, V.; Debiak, P.; Dennison, C.;
  Farhi, D.; Fischer, Q.; Hashme, S.; Hesse, C.; et~al. 2019.
\newblock Dota 2 with large scale deep reinforcement learning.
\newblock \emph{arXiv preprint arXiv:1912.06680}.

\bibitem[{Delecki, Corso, and Kochenderfer(2023)}]{delecki2023model}
Delecki, H.; Corso, A.; and Kochenderfer, M. 2023.
\newblock Model-based Validation as Probabilistic Inference.
\newblock In \emph{Learning for Dynamics and Control Conference}, 825--837.
  PMLR.

\bibitem[{Dennis et~al.(2020)Dennis, Jaques, Vinitsky, Bayen, Russell, Critch,
  and Levine}]{dennis2020emergent}
Dennis, M.; Jaques, N.; Vinitsky, E.; Bayen, A.; Russell, S.; Critch, A.; and
  Levine, S. 2020.
\newblock Emergent Complexity and Zero-shot Transfer via Unsupervised
  Environment Design.
\newblock \emph{Advances in Neural Information Processing Systems (NeurIPS)},
  33: 13049--13061.

\bibitem[{Freeman et~al.(2021)Freeman, Frey, Raichuk, Girgin, Mordatch, and
  Bachem}]{brax2021github}
Freeman, C.~D.; Frey, E.; Raichuk, A.; Girgin, S.; Mordatch, I.; and Bachem, O.
  2021.
\newblock Brax - A Differentiable Physics Engine for Large Scale Rigid Body
  Simulation.

\bibitem[{Fujimoto, Meger, and Precup(2019)}]{fujimoto2019off}
Fujimoto, S.; Meger, D.; and Precup, D. 2019.
\newblock Off-policy deep reinforcement learning without exploration.
\newblock In \emph{International Conference on Machine Learning (ICML)},
  2052--2062.

\bibitem[{Haarnoja et~al.(2018)Haarnoja, Zhou, Hartikainen, Tucker, Ha, Tan,
  Kumar, Zhu, Gupta, Abbeel et~al.}]{haarnoja2018soft}
Haarnoja, T.; Zhou, A.; Hartikainen, K.; Tucker, G.; Ha, S.; Tan, J.; Kumar,
  V.; Zhu, H.; Gupta, A.; Abbeel, P.; et~al. 2018.
\newblock Soft actor-critic algorithms and applications.
\newblock \emph{arXiv preprint arXiv:1812.05905}.

\bibitem[{Han et~al.(2020)Han, Xiong, Sun, Sun, Fang, Guo, Chen, Shi, Yu, Wu
  et~al.}]{han2020tstarbot}
Han, L.; Xiong, J.; Sun, P.; Sun, X.; Fang, M.; Guo, Q.; Chen, Q.; Shi, T.; Yu,
  H.; Wu, X.; et~al. 2020.
\newblock Tstarbot-x: An open-sourced and comprehensive study for efficient
  league training in starcraft ii full game.
\newblock \emph{arXiv preprint arXiv:2011.13729}.

\bibitem[{Han et~al.(2023)Han, Zhu, Sheng, Zhang, Li, Zhang, Zhang, Liu, Zhou,
  Zhao et~al.}]{han2023lifelike}
Han, L.; Zhu, Q.; Sheng, J.; Zhang, C.; Li, T.; Zhang, Y.; Zhang, H.; Liu, Y.;
  Zhou, C.; Zhao, R.; et~al. 2023.
\newblock Lifelike agility and play on quadrupedal robots using reinforcement
  learning and generative pre-trained models.
\newblock \emph{arXiv preprint arXiv:2308.15143}.

\bibitem[{Hu et~al.(2020)Hu, Anderson, Li, Sun, Carr, Ragan-Kelley, and
  Durand}]{hu2020difftaichi}
Hu, Y.; Anderson, L.; Li, T.-M.; Sun, Q.; Carr, N.; Ragan-Kelley, J.; and
  Durand, F. 2020.
\newblock DiffTaichi: Differentiable Programming for Physical Simulation.
\newblock In \emph{International Conference on Learning Representations
  (ICLR)}.

\bibitem[{Janner et~al.(2019)Janner, Fu, Zhang, and Levine}]{janner2019trust}
Janner, M.; Fu, J.; Zhang, M.; and Levine, S. 2019.
\newblock When to Trust Your Model: Model-Based Policy Optimization.
\newblock \emph{Advances in Neural Information Processing Systems (NeurIPS)},
  32: 12519--12530.

\bibitem[{Kaiser et~al.(2019)Kaiser, Babaeizadeh, Mi{\l}os, Osi{\'n}ski,
  Campbell, Czechowski, Erhan, Finn, Kozakowski, Levine
  et~al.}]{kaiser2019model}
Kaiser, {\L}.; Babaeizadeh, M.; Mi{\l}os, P.; Osi{\'n}ski, B.; Campbell, R.~H.;
  Czechowski, K.; Erhan, D.; Finn, C.; Kozakowski, P.; Levine, S.; et~al. 2019.
\newblock Model Based Reinforcement Learning for Atari.
\newblock In \emph{International Conference on Learning Representations
  (ICLR)}.

\bibitem[{Kakade and Langford(2002)}]{kakade2002}
Kakade, S.; and Langford, J. 2002.
\newblock Approximately optimal ap- proximate reinforcement learning.
\newblock In \emph{International Conference on Machine Learning (ICML)},
  volume~2, 267--274.

\bibitem[{Li et~al.(2023)Li, Xu, Dong, Yang, Yuan, Sun, and
  Han}]{li2023fittest}
Li, S.; Xu, J.; Dong, H.; Yang, Y.; Yuan, C.; Sun, P.; and Han, L. 2023.
\newblock The Fittest Wins: a Multi-Stage Framework Achieving New SOTA in
  ViZDoom Competition.
\newblock \emph{IEEE Transactions on Games}.

\bibitem[{Luo et~al.(2018)Luo, Xu, Li, Tian, Darrell, and
  Ma}]{luo2018algorithmic}
Luo, Y.; Xu, H.; Li, Y.; Tian, Y.; Darrell, T.; and Ma, T. 2018.
\newblock Algorithmic framework for model-based deep reinforcement learning
  with theoretical guarantees.
\newblock \emph{arXiv preprint arXiv:1807.03858}.

\bibitem[{Lyu et~al.(2022)Lyu, Ma, Li, and Lu}]{lyu2022mildly}
Lyu, J.; Ma, X.; Li, X.; and Lu, Z. 2022.
\newblock Mildly conservative Q-learning for offline reinforcement learning.
\newblock \emph{Advances in Neural Information Processing Systems}, 35:
  1711--1724.

\bibitem[{Mnih et~al.(2015)Mnih, Kavukcuoglu, Silver, Rusu, Veness, Bellemare,
  Graves, Riedmiller, Fidjeland, and Ostrovski}]{mnih2015human}
Mnih, V.; Kavukcuoglu, K.; Silver, D.; Rusu, A.~A.; Veness, J.; Bellemare,
  M.~G.; Graves, A.; Riedmiller, M.; Fidjeland, A.~K.; and Ostrovski, G. 2015.
\newblock Human-level control through deep reinforcement learning.
\newblock \emph{Nature}, 518(7540): 529.

\bibitem[{Rosete-Beas et~al.(2023)Rosete-Beas, Mees, Kalweit, Boedecker, and
  Burgard}]{rosete2023latent}
Rosete-Beas, E.; Mees, O.; Kalweit, G.; Boedecker, J.; and Burgard, W. 2023.
\newblock Latent plans for task-agnostic offline reinforcement learning.
\newblock In \emph{Conference on Robot Learning}, 1838--1849. PMLR.

\bibitem[{Schaefer, Udluft, and Zimmermann(2007)}]{schaefer2007recurrent}
Schaefer, A.~M.; Udluft, S.; and Zimmermann, H.-G. 2007.
\newblock A recurrent control neural network for data efficient reinforcement
  learning.
\newblock In \emph{2007 IEEE International Symposium on Approximate Dynamic
  Programming and Reinforcement Learning}, 151--157. IEEE.

\bibitem[{Schulman et~al.(2015)Schulman, Levine, Abbeel, Jordan, and
  Moritz}]{schulman2015trust}
Schulman, J.; Levine, S.; Abbeel, P.; Jordan, M.; and Moritz, P. 2015.
\newblock Trust region policy optimization.
\newblock In \emph{International Conference on Machine Learning (ICML)},
  1889--1897.

\bibitem[{Schulman et~al.(2017)Schulman, Wolski, Dhariwal, Radford, and
  Klimov}]{schulman2017proximal}
Schulman, J.; Wolski, F.; Dhariwal, P.; Radford, A.; and Klimov, O. 2017.
\newblock Proximal policy optimization algorithms.
\newblock \emph{arXiv preprint arXiv:1707.06347}.

\bibitem[{Siegel et~al.(2020)Siegel, Springenberg, Berkenkamp, Abdolmaleki,
  Neunert, Lampe, Hafner, Heess, and Riedmiller}]{siegel2020keep}
Siegel, N.~Y.; Springenberg, J.~T.; Berkenkamp, F.; Abdolmaleki, A.; Neunert,
  M.; Lampe, T.; Hafner, R.; Heess, N.; and Riedmiller, M. 2020.
\newblock Keep doing what worked: Behavioral modelling priors for offline
  reinforcement learning.
\newblock \emph{arXiv preprint arXiv:2002.08396}.

\bibitem[{Silver et~al.(2016)Silver, Huang, Maddison, Guez, Sifre, Van
  Den~Driessche, Schrittwieser, Antonoglou, Panneershelvam, and
  Lanctot}]{silver2016mastering}
Silver, D.; Huang, A.; Maddison, C.~J.; Guez, A.; Sifre, L.; Van Den~Driessche,
  G.; Schrittwieser, J.; Antonoglou, I.; Panneershelvam, V.; and Lanctot, M.
  2016.
\newblock Mastering the game of Go with deep neural networks and tree search.
\newblock \emph{Nature}, 529(7587): 484.

\bibitem[{Silver et~al.(2017)Silver, Schrittwieser, Simonyan, Antonoglou,
  Huang, Guez, Hubert, Baker, Lai, and Bolton}]{silver2017mastering}
Silver, D.; Schrittwieser, J.; Simonyan, K.; Antonoglou, I.; Huang, A.; Guez,
  A.; Hubert, T.; Baker, L.; Lai, M.; and Bolton, A. 2017.
\newblock Mastering the game of Go without human knowledge.
\newblock \emph{Nature}, 550(7676): 354.

\bibitem[{Taylor and Stone(2009)}]{taylor2009transfer}
Taylor, M.~E.; and Stone, P. 2009.
\newblock Transfer learning for reinforcement learning domains: A survey.
\newblock \emph{Journal of Machine Learning Research}, 10(7).

\bibitem[{Vinyals et~al.(2019)Vinyals, Babuschkin, Czarnecki, Mathieu, Dudzik,
  Chung, Choi, Powell, Ewalds, Georgiev et~al.}]{astar}
Vinyals, O.; Babuschkin, I.; Czarnecki, W.~M.; Mathieu, M.; Dudzik, A.; Chung,
  J.; Choi, D.~H.; Powell, R.; Ewalds, T.; Georgiev, P.; et~al. 2019.
\newblock Grandmaster Level in StarCraft II using Multi-Agent Reinforcement
  Learning.
\newblock \emph{Nature}, 575(7782): 350--354.

\bibitem[{Wang et~al.(2023)Wang, Wongkamjan, Jia, and Huang}]{wang2023live}
Wang, X.; Wongkamjan, W.; Jia, R.; and Huang, F. 2023.
\newblock Live in the moment: Learning dynamics model adapted to evolving
  policy.
\newblock In \emph{International Conference on Machine Learning}, 36470--36493.
  PMLR.

\bibitem[{Werling et~al.(2021)Werling, Omens, Lee, Exarchos, and
  Liu}]{werling2021fast}
Werling, K.; Omens, D.; Lee, J.; Exarchos, I.; and Liu, C.~K. 2021.
\newblock Fast and Feature-Complete Differentiable Physics for Articulated
  Rigid Bodies with Contact.
\newblock \emph{arXiv preprint arXiv:2103.16021}.

\bibitem[{Wu et~al.(2022)Wu, Yu, Chen, Hao, and Zhuo}]{wu2022plan}
Wu, Z.; Yu, C.; Chen, C.; Hao, J.; and Zhuo, H.~H. 2022.
\newblock Plan To Predict: Learning an Uncertainty-Foreseeing Model For
  Model-Based Reinforcement Learning.
\newblock \emph{Advances in Neural Information Processing Systems}, 35:
  15849--15861.

\bibitem[{Xu et~al.(2023)Xu, Li, Yang, Yuan, and Han}]{xu2023efficient}
Xu, J.; Li, S.; Yang, R.; Yuan, C.; and Han, L. 2023.
\newblock Efficient Multi-Goal Reinforcement Learning via Value Consistency
  Prioritization.
\newblock \emph{Journal of Artificial Intelligence Research}, 77: 355--376.

\bibitem[{Zhan et~al.(2022)Zhan, Huang, Huang, Jiang, and
  Lee}]{zhan2022offline}
Zhan, W.; Huang, B.; Huang, A.; Jiang, N.; and Lee, J. 2022.
\newblock Offline reinforcement learning with realizability and single-policy
  concentrability.
\newblock In \emph{Conference on Learning Theory}, 2730--2775. PMLR.

\bibitem[{Zhu et~al.(2023)Zhu, Lin, Jain, and Zhou}]{zhu2023transfer}
Zhu, Z.; Lin, K.; Jain, A.~K.; and Zhou, J. 2023.
\newblock Transfer learning in deep reinforcement learning: A survey.
\newblock \emph{IEEE Transactions on Pattern Analysis and Machine
  Intelligence}.

\end{thebibliography}

\appendix
\onecolumn

\section{Proof of The Relativity Lemma~\ref{theo:main}}
\label{sec:proof_relativity}
\proof
Define the expected cumulative return as
{\small
\[
J(\P,\pi)=\E_{s_0,a_0,\cdots\sim\P,\pi}\left[
\sum_{t=0}^{\infty}\gamma^t r(s_t,a_t,s_{t+1})
\right],
\]
}\noindent
where
{\small
\[
s_0\sim\rho(\P_0),\ a_t\sim\pi(a_t|s_t),\ s_{t+1}\sim\P(s_{t+1}|s_t,a_t).
\]
}\noindent
The expected cumulative return can be expanded as
{\small
\begin{align*}
J(\P,\pi)
&=\sum_{t=0}^{\infty}\gamma^t\E_{s_t,a_t,s_{t+1}\sim\P,\pi}
r(s_t,a_t,s_{t+1})
\\
&=\sum_{t=0}^{\infty}\gamma^t\E_{s_t\sim p^{\P,\pi}(s_t),a_t\sim\pi(a_t|s_t),s_{t+1}\sim\P(s_{t+1}|s_t,a_t)}
r(s_t,a_t,s_{t+1})
\\
&=\sum_{t=0}^{\infty}\gamma^t\sum_{s_t}p^{\P,\pi}(s_t)
\sum_{a_t}\pi(a_t|s_t)
\sum_{s_{t+1}}\P(s_{t+1}|s_t,a_t)
r(s_t,a_t,s_{t+1}),
\end{align*}
}\noindent
where $p^{\P,\pi}(s_t)$ indicates the marginal probability that $s_t$ is visited under $\P$ and $\pi$.
With a fixed policy $\pi$, we investigate the value difference between two MDPs as
{\small
\begin{align}
&J(\P',\pi)-J(\P,\pi)
\nonumber
\\
=&\sum_{t=0}^{\infty}\gamma^t
\big[
\E_{s_t\sim p^{\P',\pi}(s_t),a_t\sim\pi(a_t|s_t),s_{t+1}\sim\P'(s_{t+1}|s_t,a_t)}
r(s_t,a_t,s_{t+1})-
\nonumber
\\
&\qquad\quad
\E_{s_t\sim p^{\P,\pi}(s_t),a_t\sim\pi(a_t|s_t),s_{t+1}\sim\P(s_{t+1}|s_t,a_t)}
r(s_t,a_t,s_{t+1})
\big]
\nonumber
\\
=&\sum_{t=0}^{\infty}\gamma^t\bigg[
\sum_{s_t}p^{\P',\pi}(s_t)
\sum_{a_t}\pi(a_t|s_t)
\sum_{s_{t+1}}\left(\P'(s_{t+1}|s_t,a_t)-\P(s_{t+1}|s_t,a_t)\right)
r(s_t,a_t,s_{t+1})+
\nonumber
\\
&\qquad\quad
\sum_{s_t}\left(p^{\P',\pi}(s_t)-p^{\P,\pi}(s_t)\right)
\sum_{a_t}\pi(a_t|s_t)
\sum_{s_{t+1}}\P(s_{t+1}|s_t,a_t)
r(s_t,a_t,s_{t+1})\bigg].
\label{eq:1}
\end{align}
}\noindent

The first term in Eq.~(\ref{eq:1}) derives as
{\small
\begin{align*}
\ &\sum_{s_t}p^{\P',\pi}(s_t)
\sum_{a_t}\pi(a_t|s_t)
\sum_{s_{t+1}}\left(\P'(s_{t+1}|s_t,a_t)-\P(s_{t+1}|s_t,a_t)\right)
r(s_t,a_t,s_{t+1})
\\
=\ &\sum_{s_t}\frac{p^{\P',\pi}(s_t)}{p^{\P,\pi}(s_t)}p^{\P,\pi}(s_t)
\sum_{a_t}\pi(a_t|s_t)
\sum_{s_{t+1}}\frac{\left(\P'(s_{t+1}|s_t,a_t)-\P(s_{t+1}|s_t,a_t)\right)}{\P(s_{t+1}|s_t,a_t)}\P(s_{t+1}|s_t,a_t)
r(s_t,a_t,s_{t+1})
\\
=\ &\E_{s_t\sim p^{\P,\pi}(s_t),a_t\sim\pi(a_t|s_t),s_{t+1}\sim\P(s_{t+1}|s_t,a_t)}
\frac{p^{\P',\pi}(s_t)}{p^{\P,\pi}(s_t)}\frac{\P'(s_{t+1}|s_t,a_t)-\P(s_{t+1}|s_t,a_t)}{\P(s_{t+1}|s_t,a_t)}r(s_t,a_t,s_{t+1})
\\
=\ &\E_{s_t,a_t,s_{t+1}\sim \P,\pi}
\frac{p^{\P',\pi}(s_t)}{p^{\P,\pi}(s_t)}\frac{\P'(s_{t+1}|s_t,a_t)-\P(s_{t+1}|s_t,a_t)}{\P(s_{t+1}|s_t,a_t)}r(s_t,a_t,s_{t+1}),
\end{align*}
}\noindent
where we briefly write $s_t,a_t,s_{t+1}\sim \P,\pi$ to represent $s_t\sim p^{\P,\pi}(s_t),a_t\sim\pi(a_t|s_t),s_{t+1}\sim\P(s_{t+1}|s_t,a_t)$, when it is clear from the context around. In the second term of Eq.~(\ref{eq:1}), the difference of the marginal probabilities $p^{\P',\pi}(s_t)-p^{\P,\pi}(s_t)$ can be expanded similarly as
{\small
\begin{align}
&p^{\P',\pi}(s_t)-p^{\P,\pi}(s_t)
\nonumber
\\
=\ &\E_{s_{t-1}\sim p^{\P',\pi}(s_{t-1}),a_{t-1}\sim\pi(a_{t-1}|s_{t-1})}\ \P'(s_t|s_{t-1},a_{t-1})-
\nonumber
\\
\ &\E_{s_{t-1}\sim p^{\P,\pi}(s_{t-1}),a_{t-1}\sim\pi(a_{t-1}|s_{t-1})}\ \P(s_t|s_{t-1},a_{t-1})
\nonumber
\\
=\ &\sum_{s_{t-1}}p^{\P',\pi}(s_{t-1})\sum_{a_{t-1}}\pi(a_{t-1}|s_{t-1})
\left(\P'(s_t|s_{t-1},a_{t-1})-\P(s_t|s_{t-1},a_{t-1})\right)+
\nonumber
\\
\ &\sum_{s_{t-1}}\left(p^{\P',\pi}(s_{t-1})-p^{\P,\pi}(s_{t-1})\right)\sum_{a_{t-1}}\pi(a_{t-1}|s_{t-1})
\P(s_t|s_{t-1},a_{t-1})
\nonumber
\\
=\ &\E_{s_{t-1}\sim p^{\P,\pi}(s_{t-1}),a_{t-1}\sim\pi(a_{t-1}|s_{t-1})}
\frac{p^{\P',\pi}(s_{t-1})}{p^{\P,\pi}(s_{t-1})}\left(\P'(s_t|s_{t-1},a_{t-1})-\P(s_t|s_{t-1},a_{t-1})\right)+
\nonumber
\\
\ &\sum_{s_{t-1}}\left(p^{\P',\pi}(s_{t-1})-p^{\P,\pi}(s_{t-1})\right)\E_{a_{t-1}\sim\pi(a_{t-1}|s_{t-1})}\P(s_t|s_{t-1},a_{t-1}).
\label{eq:2}
\end{align}
}\noindent

Plugging Eq.~(\ref{eq:2}) back into the corresponding term of Eq.~(\ref{eq:1}), we have
{\small
\begin{align}
&\sum_{s_t}\left(p^{\P',\pi}(s_t)-p^{\P,\pi}(s_t)\right)
\sum_{a_t}\pi(a_t|s_t)
\sum_{s_{t+1}}\P(s_{t+1}|s_t,a_t)
r(s_t,a_t,s_{t+1})
\nonumber
\\
=\ &\sum_{s_t}\left(p^{\P',\pi}(s_t)-p^{\P,\pi}(s_t)\right)
\E_{a_t\sim\pi(a_t|s_t),s_{t+1}\sim\P(s_{t+1}|s_t,a_t)}
r(s_t,a_t,s_{t+1})
\nonumber
\\
=\ &\E_{s_{t-1},a_{t-1},s_t,a_t,s_{t+1}\sim\P,\pi}
\frac{p^{\P',\pi}(s_{t-1})}{p^{\P,\pi}(s_{t-1})}
\frac{\P'(s_t|s_{t-1},a_{t-1})-\P(s_t|s_{t-1},a_{t-1})}{\P(s_t|s_{t-1},a_{t-1})}
r(s_t,a_t,s_{t+1})+
\nonumber
\\
&\sum_{s_{t-1}}\left(p^{\P',\pi}(s_{t-1})-p^{\P,\pi}(s_{t-1})\right)\E_{a_{t-1},s_t,a_t,s_{t+1}\sim\P,\pi}
r(s_t,a_t,s_{t+1}).
\label{eq:3}
\end{align}
}\noindent

Note that the second term in Eq.~(\ref{eq:3}) has an identical form as that in Eq.~(\ref{eq:1}) by expanding prior states and actions. Therefore, by recursively expanding Eq.~(\ref{eq:3}) backward, we finally have
{\small
\begin{align}
&J(\P',\pi)-J(\P,\pi)
\nonumber
\\
=\ &\sum_{t=0}^{\infty}\gamma^t\bigg[\E_{s_t,a_t,s_{t+1}\sim\P,\pi}
\frac{p^{\P',\pi}(s_t)}{p^{\P,\pi}(s_t)}\frac{\P'(s_{t+1}|s_t,a_t)-\P(s_{t+1}|s_t,a_t)}{\P(s_{t+1}|s_t,a_t)}r(s_t,a_t,s_{t+1})+
\nonumber
\\
&\ \E_{s_{t-1},a_{t-1},s_t,a_t,s_{t+1}\sim\P,\pi}
\frac{p^{\P',\pi}(s_{t-1})}{p^{\P,\pi}(s_{t-1})}
\frac{\P'(s_t|s_{t-1},a_{t-1})-\P(s_t|s_{t-1},a_{t-1})}{\P(s_t|s_{t-1},a_{t-1})}
r(s_t,a_t,s_{t+1})+
\nonumber
\\
&\ \cdots\cdots\bigg]
\nonumber
\\
=\ &\sum_{t=0}^{\infty}\E_{\tau\sim\P,\pi}\ 
\sum_{i=0}^t\gamma^t
\frac{p^{\P',\pi}(s_i)}{p^{\P,\pi}(s_i)}
\frac{\P'(s_{i+1}|s_i,a_i)-\P(s_{i+1}|s_i,a_i)}{\P(s_{i+1}|s_i,a_i)}
r(s_t,a_t,s_{t+1})
\nonumber
\\
=\ &\E_{\tau\sim\P,\pi}\ 
\sum_{t=0}^{\infty}
\gamma^t r(s_t,a_t,s_{t+1})
\sum_{i=0}^t
\frac{p^{\P',\pi}(s_i)}{p^{\P,\pi}(s_i)}
\frac{\P'(s_{i+1}|s_i,a_i)-\P(s_{i+1}|s_i,a_i)}{\P(s_{i+1}|s_i,a_i)},
\label{eq:4}
\end{align}
}\noindent
where $\tau$ indicates the trajectory $s_0,a_0,\cdots$.

Let $x_t=\gamma^t r(s_t,a_t,s_{t+1})$ and $y_i=\frac{p^{\P',\pi}(s_i)}{p^{\P,\pi}(s_i)}
\frac{\P'(s_{i+1}|s_i,a_i)-\P(s_{i+1}|s_i,a_i)}{\P(s_{i+1}|s_i,a_i)}$, we have
{\small
\begin{align*}
\sum_{t=0}^{\infty}x_t\sum_{i=0}^t y_i
&=x_0y_0+x_1(y_0+y_1)+x_2(y_0+y_1+y_2)+\cdots
\\
&=y_0(x_0+x_1+x_2+\cdots)+y_1(x_1+x_2+\cdots)+\cdots
=\sum_{i=0}^{\infty}y_i\sum_{t=i}^{\infty}x_t,
\end{align*}
}\noindent
and $\sum_{t=i}^\infty x_t=\gamma^{i}\sum_{t=i}^\infty\gamma^{t-i} r(s_t,a_t,s_{t+1})=\gamma^i R_i$, where $R_i=\sum_{t=i}^\infty\gamma^{t-i} r(s_t,a_t,s_{t+1})$ is the empirical cumulative future reward. Continuing from Eq.~(\ref{eq:4}),
{\small
\begin{align}
&J(\P',\pi)-J(\P,\pi)
\nonumber
\\
=\ &
\E_{\tau\sim\P,\pi}\sum_{t=0}^{\infty}\gamma^t R_t
\frac{p^{\P',\pi}(s_t)}{p^{\P,\pi}(s_t)}
\frac{\P'(s_{t+1}|s_t,a_t)-\P(s_{t+1}|s_t,a_t)}{\P(s_{t+1}|s_t,a_t)}
\nonumber
\\
=\ &
\sum_{t=0}^{\infty}\gamma^t
\E_{\tau\sim\P,\pi}
R_t\frac{p^{\P',\pi}(s_t)}{p^{\P,\pi}(s_t)}
\frac{\P'(s_{t+1}|s_t,a_t)-\P(s_{t+1}|s_t,a_t)}{\P(s_{t+1}|s_t,a_t)}
\nonumber
\\
=\ &\sum_{t=0}^{\infty}\gamma^t
\E_{s_t,a_t,s_{t+1}\sim\P,\pi}\ 
\frac{p^{\P',\pi}(s_t)}{p^{\P,\pi}(s_t)}
\frac{\P'(s_{t+1}|s_t,a_t)-\P(s_{t+1}|s_t,a_t)}{\P(s_{t+1}|s_t,a_t)}
\E_{a_{t+1},s_{t+2},\cdots\sim\P,\pi}\ R_t
\nonumber
\\
=\ &\sum_{t=0}^{\infty}\gamma^t
\E_{s_t,a_t,s_{t+1}\sim\P,\pi}\ 
\frac{p^{\P',\pi}(s_t)}{p^{\P,\pi}(s_t)}
\frac{\P'(s_{t+1}|s_t,a_t)-\P(s_{t+1}|s_t,a_t)}{\P(s_{t+1}|s_t,a_t)}
\E_{a_{t+1},s_{t+2},\cdots\sim\P,\pi}
\left[
r(s_t,a_t,s_{t+1})+\gamma R_{t+1}
\right]
\nonumber
\\
=\ &\sum_{t=0}^{\infty}\gamma^t
\E_{s_t,a_t,s_{t+1}\sim\P,\pi}\ 
\frac{p^{\P',\pi}(s_t)}{p^{\P,\pi}(s_t)}
\frac{\P'(s_{t+1}|s_t,a_t)-\P(s_{t+1}|s_t,a_t)}{\P(s_{t+1}|s_t,a_t)}
\left(
r(s_t,a_t,s_{t+1})+\gamma V^{\P,\pi}(s_{t+1})
\right)
\nonumber
\\
=\ &\sum_{t=0}^{\infty}\gamma^t
\E_{s_t,a_t\sim\P',\pi}\E_{s_{t+1}\sim\P(s_{t+1}|s_t,a_t)}
\frac{\P'(s_{t+1}|s_t,a_t)-\P(s_{t+1}|s_t,a_t)}{\P(s_{t+1}|s_t,a_t)}
\left(
r(s_t,a_t,s_{t+1})+\gamma V^{\P,\pi}(s_{t+1})
\right)
\nonumber
\\
=\ &\sum_{t=0}^{\infty}\gamma^t\left\{
\E_{s_t,a_t,s_{t+1}\sim\P',\pi}
\big[
r(s_t,a_t,s_{t+1})+\gamma V^{\P,\pi}(s_{t+1})
\big]
-\E_{s_t,a_t\sim\P',\pi}\ Q^{\P,\pi}(s_t,a_t)
\right\}
\nonumber
\\
=\ &\E_{\tau\sim\P',\pi}\sum_{t=0}^{\infty}\gamma^t\ 
\left[
r(s_t,a_t,s_{t+1})+\gamma V^{\P,\pi}(s_{t+1})-Q^{\P,\pi}(s_t,a_t)
\right].
\label{eq:diff}
\end{align}
}\noindent
Now, we complete the proof.

\rightline{$\Box$}

\section{Some Useful Lemmas}
\begin{lemma}
\label{proof:lemma1}
Let $p^{\P,\pi}(s)$ indicate the marginal visitation probability for state $s$ under dynamics $\P$ and policy $\pi$. Let $\delta_1$ and $\delta$ be the total variation divergence of the dynamics and policies, respectively, as defined in Theorem~\ref{theo:2}. With the assumption used in Lemma~\ref{theo:main} that $\P'(s_0)=\P(s_0)$, we have
{\small
\begin{align}
&\sum_{s}|p^{\P',\pi}(s)-p^{\P,\pi}(s)|\leq 2t\delta_1,
\label{lemma:pd_diff}
\\
&\sum_{s}|p^{\P',\pi}(s)-p^{\P,\pi}(s)|\leq 2t\delta_2.
\label{lemma:pp_diff}
\end{align}
}\noindent
\proof
From Eq.~\eqref{eq:2}, we know that
{\small
\begin{align*}
&p^{\P',\pi}(s_t)-p^{\P,\pi}(s_t)
\\
=\ &\sum_{s_{t-1}}p^{\P',\pi}(s_{t-1})\sum_{a_{t-1}}\pi(a_{t-1}|s_{t-1})
\left(\P'(s_t|s_{t-1},a_{t-1})-\P(s_t|s_{t-1},a_{t-1})\right)+
\\
\ &\sum_{s_{t-1}}\underbrace{
\left(p^{\P',\pi}(s_{t-1})-p^{\P,\pi}(s_{t-1})\right)
}_{\text{probability discrepancy at time step t-1}}
\sum_{a_{t-1}}\pi(a_{t-1}|s_{t-1})
\P(s_t|s_{t-1},a_{t-1}).
\end{align*}
}\noindent
Recursively expanding the above equation will get
{\small
\begin{align*}
&p^{\P',\pi}(s_t)-p^{\P,\pi}(s_t)
\\
=\ &\sum_{s_{t-1}}p^{\P',\pi}(s_{t-1})\sum_{a_{t-1}}\pi(a_{t-1}|s_{t-1})
\left(\P'(s_t|s_{t-1},a_{t-1})-\P(s_t|s_{t-1},a_{t-1})\right)+
\\
\ &\sum_{s_{t-1}}\sum_{s_{t-2}}p^{\P',\pi}(s_{t-2})\sum_{a_{t-2}}\pi(a_{t-2}|s_{t-2})
\left(\P'(s_{t-1}|s_{t-2},a_{t-2})-\P(s_{t-1}|s_{t-2},a_{t-2})\right)
\sum_{a_{t-1}}\pi(a_{t-1}|s_{t-1})
\P(s_t|s_{t-1},a_{t-1})
+
\\
\ &\sum_{s_{t-1}}\sum_{s_{t-2}}
\underbrace{\left(p^{\P',\pi}(s_{t-2})-p^{\P,\pi}(s_{t-2})\right)}_{\text{probability discrepancy at time step t-2}}
\sum_{a_{t-2}}\pi(a_{t-2}|s_{t-2})
\P(s_{t-1}|s_{t-2},a_{t-2})
\sum_{a_{t-1}}\pi(a_{t-1}|s_{t-1})
\P(s_t|s_{t-1},a_{t-1})
\\
=\ &\cdots
\end{align*}
}\noindent
By noting that $p^{\P',\pi}(s_0)=\P'(s_0)=\P(s_0)=p^{\P,\pi}(s_0)$, the last expanded term containing
$(p^{\P',\pi}(s_0)-p^{\P,\pi}(s_0))$ equals zero, and
{\small
\[
\sum_{s_i}|\P'(s_i|s_{i-1},a_{i-1})-\P(s_i|s_{i-1},a_{i-1})|\leq 2\delta_1, \text{for any $i$}.
\] 
}\noindent
Then, it is easy to obtain
{\small
\begin{align*}
\sum_{s_t}|p^{\P',\pi}(s_t)-p^{\P,\pi}(s_t)|\leq2t\delta_1.
\end{align*}
}\noindent
Similarly, we have
{\small
\begin{align*}
&p^{\P,\pi'}(s_t)-p^{\P,\pi}(s_t)
\\
=\ &\sum_{s_{t-1}}p^{\P,\pi'}(s_{t-1})\sum_{a_{t-1}}\pi'(a_{t-1}|s_{t-1})\P(s_t|s_{t-1},a_{t-1})-
\sum_{s_{t-1}}p^{\P,\pi}(s_{t-1})\sum_{a_{t-1}}\pi(a_{t-1}|s_{t-1})\P(s_t|s_{t-1},a_{t-1})
\\
=\ &\sum_{s_{t-1}}p^{\P,\pi'}(s_{t-1})\sum_{a_{t-1}}\pi'(a_{t-1}|s_{t-1})\P(s_t|s_{t-1},a_{t-1})-
\sum_{s_{t-1}}p^{\P,\pi}(s_{t-1})\sum_{a_{t-1}}\pi'(a_{t-1}|s_{t-1})\P(s_t|s_{t-1},a_{t-1})+
\\
\ &\sum_{s_{t-1}}p^{\P,\pi}(s_{t-1})\sum_{a_{t-1}}\pi'(a_{t-1}|s_{t-1})\P(s_t|s_{t-1},a_{t-1})-
\sum_{s_{t-1}}p^{\P,\pi}(s_{t-1})\sum_{a_{t-1}}\pi(a_{t-1}|s_{t-1})\P(s_t|s_{t-1},a_{t-1})
\\
=\ &\sum_{s_{t-1}}
\underbrace{\left(p^{\P,\pi'}(s_{t-1})-p^{\P,\pi}(s_{t-1})\right)
}_{\text{probability discrepancy at time step t-1}}
\sum_{a_{t-1}}\pi'(a_{t-1}|s_{t-1})\P(s_t|s_{t-1},a_{t-1})+
\\
\ &\sum_{s_{t-1}}p^{\P,\pi}(s_{t-1})\sum_{a_{t-1}}\left(\pi'(a_{t-1}|s_{t-1})-\pi(a_{t-1}|s_{t-1})\right)\P(s_t|s_{t-1},a_{t-1})
\\
=\ &\sum_{s_{t-1}}\sum_{s_{t-2}}
\underbrace{\left(p^{\P,\pi'}(s_{t-2})-p^{\P,\pi}(s_{t-2})\right)
}_{\text{probability discrepancy at time step t-2}}
\sum_{a_{t-2}}\pi'(a_{t-2}|s_{t-2})\P(s_{t-1}|s_{t-2},a_{t-2})
\sum_{a_{t-1}}\pi'(a_{t-1}|s_{t-1})\P(s_t|s_{t-1},a_{t-1})+
\\
\ &\sum_{s_{t-1}}\sum_{s_{t-2}}p^{\P,\pi}(s_{t-2})\sum_{a_{t-2}}\left(\pi'(a_{t-2}|s_{t-2})-\pi(a_{t-2}|s_{t-2})\right)\P(s_{t-1}|s_{t-2},a_{t-2})
\sum_{a_{t-1}}\pi'(a_{t-1}|s_{t-1})\P(s_t|s_{t-1},a_{t-1})+
\\
\ &\sum_{s_{t-1}}p^{\P,\pi}(s_{t-1})\sum_{a_{t-1}}\left(\pi'(a_{t-1}|s_{t-1})-\pi(a_{t-1}|s_{t-1})\right)\P(s_t|s_{t-1},a_{t-1})
\\
=\ &\cdots
\end{align*}
}\noindent
By noting that $p^{\P,\pi'}(s_0)=\P(s_0)=p^{\P,\pi}(s_0)$, the last expanded term containing
$(p^{\P,\pi'}(s_0)-p^{\P,\pi}(s_0))$ equals zero, and
{\small
\[
\sum_{a_i}|\pi'(a_i|s_i)-\pi(a_i|s_i)|\leq 2\delta_1, \text{for any $i$}.
\]
}\noindent
Then, it is easy to obtain
{\small
\begin{align*}
\sum_{s_t}|p^{\P,\pi'}(s_t)-p^{\P,\pi}(s_t)|\leq2t\delta_2.
\end{align*}
}\noindent
\rightline{$\Box$}
\end{lemma}

\begin{lemma}
\label{proof:lemma2}
With the definitions in Theorem~\ref{theo:2} and Lemma~\ref{proof:lemma1}, we have the following bounds on the value differences at a single state $s_t$.
{\small
\begin{align}
&\left|V^{\P',\pi}(s_t)-V^{\P,\pi}(s_t)\right|\leq
\min\left(2r_{max}\delta_1\frac{t\gamma(1-\gamma)+1}{(1-\gamma)^2}, 
\frac{2r_{max}}{1-\gamma}\right),
\label{eq:vd_diff}
\\
&\left|V^{\P,\pi'}(s_t)-V^{\P,\pi}(s_t)\right|\leq
\min\left(2r_{max}\delta_2\frac{t\gamma(1-\gamma)+1}{(1-\gamma)^2}, \frac{2r_{max}}{1-\gamma}\right).
\label{eq:vp_diff}
\end{align}
}\noindent
\proof
First, it is easy to see that for any $\P$, $\pi$ and state $s_t$ we always have
{\small
\[
\left|V^{\P,\pi}(s_t)\right|=\E_{a_t,s_{t+1},\cdots\sim\P',\pi}\sum_{i=t}^{\infty}\gamma^{i-t}r(s_i,a_i,s_{i+1})
\leq\frac{r_{max}}{1-\gamma},
\]
}\noindent
and hence
{\small
\[
\left|V^{\P',\pi}(s_t)-V^{\P,\pi}(s_t)\right|
\leq\frac{2r_{max}}{1-\gamma}.
\]
}\noindent
Next, we derive another bound for the value differences. We have
{\small
\begin{align*}
&V^{\P',\pi}(s_t)-V^{\P,\pi}(s_t)
\\
=\ &\E_{a_t,s_{t+1},\cdots\sim\P',\pi}\sum_{i=t}^{\infty}\gamma^{i-t}r(s_i,a_i,s_{i+1})-
\E_{a_t,s_{t+1},\cdots\sim\P,\pi}\sum_{i=t}^{\infty}\gamma^{i-t}r(s_i,a_i,s_{i+1})
\\
=\ &\E_{a_t,s_{t+1}\sim\P',\pi}r(s_t,a_t,s_{t+1})-\E_{a_t,s_{t+1}\sim\P,\pi}r(s_t,a_t,s_{t+1})+
\\
\ &\sum_{i=t+1}^{\infty}\gamma^{i-t}\E_{s_i,a_i,s_{i+1}\sim\P',\pi}r(s_i,a_i,s_{i+1})-
\sum_{i=t+1}^{\infty}\gamma^{i-t}\E_{s_i,a_i,s_{i+1}\sim\P,\pi}r(s_i,a_i,s_{i+1})
\\
=\ &\sum_{a_t}\pi(a_t|s_t)
\sum_{s_{t+1}}\left(\P'(s_{t+1}|s_t,a_t)-\P(s_{t+1}|s_t,a_t)\right)r(s_t,a_t,s_{t+1})+
\\
\ &\sum_{i=t+1}^{\infty}\gamma^{i-t}\sum_{s_i}
\left(p^{\P',\pi}(s_i)-p^{\P,\pi}(s_i)\right)
\sum_{a_i}\pi(a_i|s_i)
\sum_{s_{i+1}}\P'(s_{i+1}|s_i,a_i)
r(s_i,a_i,s_{i+1})+
\\
\ &\sum_{i=t+1}^{\infty}\gamma^{i-t}\sum_{s_i}p^{\P,\pi}(s_i)
\sum_{a_i}\pi(a_i|s_i)
\sum_{s_{i+1}}\left(\P'(s_{i+1}|s_i,a_i)-\P(s_{i+1}|s_i,a_i)\right)
r(s_i,a_i,s_{i+1}).
\end{align*}
}\noindent
Therefore, applying Lemma~\ref{proof:lemma1}, we have
{\small
\begin{align*}
\left|V^{\P',\pi}(s_t)-V^{\P,\pi}(s_t)\right|
\leq\ &
2r_{max}\delta_1+
r_{max}\sum_{i=t+1}^{\infty}\gamma^{i-t}\sum_{s_i}\left|p^{\P',\pi}(s_i)-p^{\P,\pi}(s_i)\right|+
2r_{max}\delta_1\sum_{i=t+1}^{\infty}\gamma^{i-t}
\\
\leq\ &2r_{max}\delta_1+2r_{max}\delta_1\sum_{i=t+1}^{\infty}i\gamma^{i-t}
+2r_{max}\delta_1\sum_{i=t+1}^{\infty}\gamma^{i-t}
\\
=\ &
2r_{max}\delta_1\frac{t\gamma(1-\gamma)+1}{(1-\gamma)^2}.
\end{align*}
}\noindent

Similarly, we have
{\small
\begin{align*}
&V^{\P,\pi'}(s_t)-V^{\P,\pi}(s_t)
\\
=\ &\E_{a_t,s_{t+1},\cdots\sim\P,\pi'}\sum_{i=t}^{\infty}\gamma^{i-t}r(s_i,a_i,s_{i+1})-
\E_{a_t,s_{t+1},\cdots\sim\P,\pi}\sum_{i=t}^{\infty}\gamma^{i-t}r(s_i,a_i,s_{i+1})
\\
=\ &\E_{a_t,s_{t+1}\sim\P,\pi'}r(s_t,a_t,s_{t+1})-\E_{a_t,s_{t+1}\sim\P,\pi}r(s_t,a_t,s_{t+1})+
\\
\ &\sum_{i=t+1}^{\infty}\gamma^{i-t}\E_{s_i,a_i,s_{i+1}\sim\P,\pi'}r(s_i,a_i,s_{i+1})-
\sum_{i=t+1}^{\infty}\gamma^{i-t}\E_{s_i,a_i,s_{i+1}\sim\P,\pi}r(s_i,a_i,s_{i+1})
\\
=\ &\sum_{a_t}\left(\pi'(a_t|s_t)-\pi(a_t|s_t)\right)\sum_{s_{t+1}}\P(s_{t+1}|s_t,a_t)r(s_t,a_t,s_{t+1})+
\\
\ &\sum_{i=t+1}^{\infty}\gamma^{i-t}\sum_{s_i}p^{\P,\pi'}(s_i)
\sum_{a_i}\left(\pi'(a_i|s_i)-\pi(a_i|s_i)\right)
\sum_{s_{i+1}}\P(s_{i+1}|s_i,a_i)r(s_i,a_i,s_{i+1})+
\\
\ &\sum_{i=t+1}^{\infty}\gamma^{i-t}\sum_{s_i}
\left(p^{\P,\pi'}(s_i)-p^{\P,\pi}(s_i)\right)
\sum_{a_i}\pi(a_i|s_i)
\sum_{s_{i+1}}\P(s_{i+1}|s_i,a_i)r(s_i,a_i,s_{i+1}).
\end{align*}
}\noindent

Therefore,
{\small
\begin{align*}
\left|V^{\P,\pi'}(s_t)-V^{\P,\pi}(s_t)\right|
\leq\ &
r_{max}\sum_{a_t}\left|\pi'(a_t|s_t)-\pi(a_t|s_t)\right|+
\\
\ &
r_{max}\sum_{i=t+1}^{\infty}\gamma^{i-t}\sum_{s_i}p^{\P,\pi}(s_i)
\sum_{a_i}\left|\pi'(a_i|s_i)-\pi(a_i|s_i)\right|+
\\
\ &
r_{max}\sum_{i=t+1}^{\infty}\gamma^{i-t}
\sum_{s_i}\left|
p^{\P,\pi'}(s_i)-p^{\P,\pi}(s_i)
\right|
\\
\leq\ &
2r_{max}\delta_2+2r_{max}\delta_2\sum_{i=t}^{\infty}\gamma^{i-t}
+2r_{max}\delta_2\sum_{i=t+1}^{\infty}i\gamma^{i-t}
\\
=\ &
2r_{max}\delta_2\frac{t\gamma(1-\gamma)+1}{(1-\gamma)^2}.
\end{align*}
}\noindent
\rightline{$\Box$}
\end{lemma}

\section{Proof of Theorem~\ref{theo:2}}
\label{sec:proof_theorem1}
\proof
Denote the dynamics-induced value gap as a function
{\small
\begin{align*}
\Delta^{\P',\P}(\pi)&=J(\P',\pi)-J(\P,\pi)
\\
&=\E_{\tau\sim \P',\pi}\sum_{t=0}^{\infty}\gamma^t\ 
\left[
r(s_t,a_t,s_{t+1})+\gamma V^{\P,\pi}(s_{t+1})-Q^{\P,\pi}(s_t,a_t)
\right].
\end{align*}
}\noindent

Let
{\small
\begin{align*}
L_{\pi'}(\pi)=
&\sum_{t=0}^{\infty}\gamma^t\ 
\E_{s_0,a_0,\cdots,s_t\sim\P',\pi'}
\sum_{a_t}\pi(a_t|s_t)
\sum_{s_{t+1}}\P'(s_{t+1}|s_t,a_t)
\\
&\left[
r(s_t,a_t,s_{t+1})+\gamma V^{\P, {\pi'}}(s_{t+1})-Q^{\P, {\pi'}}(s_t,a_t)
\right]
\end{align*}
}\noindent
be the approximation of $\Delta^{\P',\P}(\pi)$ by sampling $(s_0,a_1,\cdots,s_t)$ and evaluating values using $\pi'$. 

Then, we have
{\small
\begin{align*}
&\Delta^{\P',\P}(\pi)-L_{\pi'}(\pi)
\\
=&\ 
\E_{\tau\sim\P', {\pi}}\sum_{t=0}^{\infty}\gamma^t\ 
\left[
r(s_t,a_t,s_{t+1})+\gamma V^{\P, {\pi}}(s_{t+1})-Q^{\P, {\pi}}(s_t,a_t)
\right]-
\\
&\ 
\E_{\tau\sim\P', {\pi}}\sum_{t=0}^{\infty}\gamma^t\ 
\left[
r(s_t,a_t,s_{t+1})+\gamma V^{\P, {\pi'}}(s_{t+1})-Q^{\P, {\pi'}}(s_t,a_t)
\right]+
\\
&\ 
\E_{\tau\sim \P', {\pi}}\sum_{t=0}^{\infty}\gamma^t\ 
\left[
r(s_t,a_t,s_{t+1})+\gamma V^{\P, {\pi'}}(s_{t+1})-Q^{\P, {\pi'}}(s_t,a_t)
\right]-
\\
&\ 
\sum_{t=0}^{\infty}\gamma^t
\E_{s_0,a_0,\cdots,s_t\sim \P', {\pi'}}
\sum_{a_t}\pi(a_t|s_t)
\sum_{s_{t+1}}\P'(s_{t+1}|s_t,a_t)
\\
&\left[
r(s_t,a_t,s_{t+1})+\gamma V^{\P, {\pi'}}(s_{t+1})-Q^{\P, {\pi'}}(s_t,a_t)
\right].
\end{align*}
}\noindent

Let
{\small
\begin{align*}
D_1=&\ 
\E_{\tau\sim \P', {\pi}}\sum_{t=0}^{\infty}\gamma^t\ 
\left[
r(s_t,a_t,s_{t+1})+\gamma V^{\P, {\pi}}(s_{t+1})-Q^{\P, {\pi}}(s_t,a_t)
\right]-
\\
&\ 
\E_{\tau\sim \P', {\pi}}\sum_{t=0}^{\infty}\gamma^t\ 
\left[
r(s_t,a_t,s_{t+1})+\gamma V^{\P, {\pi'}}(s_{t+1})-Q^{\P, {\pi'}}(s_t,a_t)
\right],
\end{align*}
}\noindent
and
{\small
\begin{align*}
D_2=&\ 
\E_{\tau\sim \P', {\pi}}\sum_{t=0}^{\infty}\gamma^t\ 
\left[
r(s_t,a_t,s_{t+1})+\gamma V^{\P, {\pi'}}(s_{t+1})-Q^{\P, {\pi'}}(s_t,a_t)
\right]-
\\
&\ 
\sum_{t=0}^{\infty}\gamma^t
\E_{s_0,a_0,\cdots,s_t\sim \P', {\pi'}}
\sum_{a_t}\pi(a_t|s_t)
\sum_{s_{t+1}}\P'(s_{t+1}|s_t,a_t)
\\
&\left[
r(s_t,a_t,s_{t+1})+\gamma V^{\P, {\pi'}}(s_{t+1})-Q^{\P, {\pi'}}(s_t,a_t)
\right].
\end{align*}
}\noindent

For $D_1$, we have
{\small
\begin{align*}
D_1
=\ &\E_{\tau\sim \P',\pi}\sum_{t=0}^{\infty}\gamma^t\ 
\left[
\gamma \bigg(V^{\P,\pi}(s_{t+1})-V^{\P,\pi'}(s_{t+1})\right)-
\\
\ &\left(
\E_{s_{t+1}\sim \P}\ 
[r(s_t,a_t,s_{t+1})+\gamma V^{\P,\pi}(s_{t+1})]-
\E_{s_{t+1}\sim \P}\ 
[r(s_t,a_t,s_{t+1})+\gamma V^{\P,\pi'}(s_{t+1})]
\bigg)
\right]
\\
=\ &
\sum_{t=0}^{\infty}\gamma^{t+1}\E_{s_0,\ldots, a_t\sim \P', \pi}\ 
\left[
\E_{s_{t+1}\sim \P'}
\bigg(V^{\P,\pi}(s_{t+1})-V^{\P,\pi'}(s_{t+1})\right)-
\\
&\E_{s_{t+1}\sim \P}
\bigg(V^{\P,\pi}(s_{t+1})-V^{\P,\pi'}(s_{t+1})\bigg)
\bigg]
\\
=\ &\sum_{t=0}^{\infty}\gamma^{t+1}\E_{s_0,\ldots, a_t\sim \P', \pi}\ 
\left[
\sum_{s_{t+1}}\left(\P'(s_{t+1}|s_t,a_t)-\P(s_{t+1}|s_t,a_t)\right)
\left(
V^{\P,\pi}(s_{t+1})-V^{\P,\pi'}(s_{t+1})
\right)
\right].
\end{align*}
}\noindent

Therefore, we have
{\small
\begin{align}
|D_1|&=\sum_{t=0}^{\infty}\gamma^{t+1}\E_{s_0,\ldots, a_t\sim \P', \pi}\ 
\left[
\sum_{s_{t+1}}\left|\P'(s_{t+1}|s_t,a_t)-\P(s_{t+1}|s_t,a_t)\right|
\left|
V^{\P,\pi}(s_{t+1})-V^{\P,\pi'}(s_{t+1})
\right|
\right]
\nonumber
\\
&\leq\sum_{t=0}^{\infty}\gamma^{t+1}\E_{s_0,\ldots, a_t\sim \P', \pi}\ 
\left[
\sum_{s_{t+1}}\left|\P'(s_{t+1}|s_t,a_t)-\P(s_{t+1}|s_t,a_t)\right|\cdot
\min\left(2r_{max}\delta_2\frac{t\gamma(1-\gamma)+1}{(1-\gamma)^2}, \frac{2r_{max}}{1-\gamma}\right)
\right]
\nonumber
\\
&\leq\sum_{t=0}^\infty\gamma^{t+1}\cdot
\min\left(4r_{max}\delta_1\delta_2\frac{t\gamma(1-\gamma)+1}{(1-\gamma)^2}, \frac{4r_{max}\delta_1}{1-\gamma}\right)
\nonumber
\\
&=
\frac{4\gamma r_{max}\delta_1}{(1-\gamma)^2}
\min\left(
\frac{\delta_2(\gamma^2+1)}{1-\gamma}, 1
\right),
\label{eq:d1}
\end{align}
}\noindent
where the first inequality in Eq.~\eqref{eq:d1} is obtained by applying Lemma~\ref{proof:lemma2}.
For $D_2$, we have
{\small
\begin{align*}
D_2=\ 
&\sum_{t=0}^{\infty}\gamma^t
\E_{s_t\sim p^{\P',\pi}, \atop {a_t\sim\pi, s_{t+1}\sim\P'}}
\left[
r(s_t,a_t,s_{t+1})+\gamma V^{\P, {\pi'}}(s_{t+1})-Q^{\P, {\pi'}}(s_t,a_t)
\right]
-
\\
&\sum_{t=0}^{\infty}\gamma^t
\E_{s_t\sim p^{\P',\pi'}, \atop {a_t\sim\pi, s_{t+1}\sim\P'}}
\left[
r(s_t,a_t,s_{t+1})+\gamma V^{\P, {\pi'}}(s_{t+1})-Q^{\P, {\pi'}}(s_t,a_t)
\right]
\\
=\ 
&\sum_{t=0}^{\infty}\gamma^t
\sum_{s_t}\big(p^{\P',\pi}(s_t)-p^{\P',\pi'}(s_t)\big)
\sum_{a_t}\pi(a_t|s_t)
\sum_{s_{t+1}}\P'(s_{t+1}|s_t,a_t)
\\
&\left[
r(s_t,a_t,s_{t+1})+\gamma V^{\P, {\pi'}}(s_{t+1})-Q^{\P, {\pi'}}(s_t,a_t)
\right]
\\
=\ 
&\sum_{t=0}^{\infty}\gamma^t
\sum_{s_t}\big(p^{\P',\pi}(s_t)-p^{\P',\pi'}(s_t)\big)
\sum_{a_t}\pi(a_t|s_t)\cdot\Bigg\{
\\
&\sum_{s_{t+1}}\P'(s_{t+1}|s_t,a_t)
\left[
r(s_t,a_t,s_{t+1})+\gamma V^{\P, {\pi'}}(s_{t+1})
\right]
-
\sum_{s_{t+1}}\P'(s_{t+1}|s_t,a_t)
Q^{\P, {\pi'}}(s_t,a_t)
\Bigg\}
\\
=\ 
&\sum_{t=0}^{\infty}\gamma^t
\sum_{s_t}\big(p^{\P',\pi}(s_t)-p^{\P',\pi'}(s_t)\big)
\sum_{a_t}\pi(a_t|s_t)\cdot\Bigg\{
\\
&\sum_{s_{t+1}}\P'(s_{t+1}|s_t,a_t)
\left[
r(s_t,a_t,s_{t+1})+\gamma V^{\P, {\pi'}}(s_{t+1})
\right]
-
Q^{\P, {\pi'}}(s_t,a_t)
\Bigg\}
\\
=\ 
&\sum_{t=0}^{\infty}\gamma^t
\sum_{s_t}\big(p^{\P',\pi}(s_t)-p^{\P',\pi'}(s_t)\big)
\sum_{a_t}\pi(a_t|s_t)
\sum_{s_{t+1}}
\\
&\left(\P'(s_{t+1}|s_t,a_t)-\P(s_{t+1}|s_t,a_t)\right)
\left(r(s_t,a_t,s_{t+1})+\gamma V^{\P, {\pi'}}(s_{t+1})\right).
\end{align*}
}\noindent

By noting that
{\small
\begin{align*}
|r(s_t,a_t,s_{t+1})+\gamma V^{\P, {\pi'}}(s_{t+1})|&=
\left|r(s_t,a_t,s_{t+1})+
\gamma\E_{a_{t+1},s_{t+2},\cdots\sim\P,\pi'}\sum_{i=t+1}^{\infty}\gamma^{i-t-1}r(s_i,a_i,s_{i+1})\right|
\leq\frac{r_{max}}{1-\gamma},
\end{align*}
}\noindent
and applying Lemma~\ref{proof:lemma1}, we have
{\small
\begin{align}
|D_2|\leq\frac{4r_{max}\delta_1\delta_2}{1-\gamma}\sum_{t=0}^\infty t\gamma^t=\frac{4\gamma r_{max}\delta_1\delta_2}{(1-\gamma)^3}.
\label{eq:d2}
\end{align}
}\noindent

Combining Eqs.~(\ref{eq:d1}) and (\ref{eq:d2}), we finally get the lower bound

{\small
\begin{align}
\Delta^{\P',\P}(\pi)
&\geq L_{\pi'}(\pi)-
\frac{4\gamma r_{max}\delta_1}{(1-\gamma)^2}
\min\left(
\frac{\delta_2(\gamma^2+1)}{1-\gamma}, 1
\right)
-
\frac{4\gamma r_{max}\delta_1\delta_2}{(1-\gamma)^3}
\nonumber
\\
&=L_{\pi'}(\pi)-
\frac{4\gamma r_{max}\delta_1}{(1-\gamma)^2}
\min\left(
\frac{\delta_2(\gamma^2+2)}{1-\gamma}, 1+\frac{\delta_2}{1-\gamma}
\right).
\label{eq:lower1}
\end{align}
}\noindent
Now, we complete the proof.

\section{Proof of Proposition~\ref{prop:1}}
\label{sec:proof_prop1}
\proof

The policy improvement theorems in TRPO~\citep{schulman2015trust} suggest
{\small
\begin{align}
J(\P,\pi)-J(\P,\pi')\geq\E_{\tau\sim\P,\pi'}
\left[
\sum_{t=0}^\infty\gamma^t A^{\P,\pi'}(s_t,a_t)
\right]-\frac{4\epsilon\gamma\delta_2^2}{(1-\gamma)^2},
\label{eq:trpo}
\end{align}
}\noindent
where $\epsilon=\max_{s,a}|A^{\P,\pi}(s,a)|$ is the maximum advantage given any $(s,a)$, defined in \citep{schulman2015trust}.
Continuing from Eq.~\eqref{eq:trpo}, we have
{\small
\begin{align}
&J(\P,\pi)-J(\P,\pi')
\nonumber
\\
\geq &\ 
\E_{\tau\sim\P,\pi'}
\left[
\sum_{t=0}^\infty\gamma^t A^{\P,\pi'}(s_t,a_t)
\right]-\frac{4\gamma\epsilon\delta_2^2}{(1-\gamma)^2}
\nonumber
\\
=&\ 
\E_{\tau\sim\P,\pi'}
\left[
\sum_{t=0}^\infty\gamma^t A^{\P,\pi'}(s_t,a_t)
\right]
+\E_{\tau\sim\P',\pi'}\left[
\sum_{t=0}^\infty\gamma^t A^{\P,\pi'}(s_t,a_t)
\right]
-\E_{\tau\sim\P',\pi'}\left[
\sum_{t=0}^\infty\gamma^t A^{\P,\pi'}(s_t,a_t)
\right]
-\frac{4\gamma\epsilon\delta_2^2}{(1-\gamma)^2}
\nonumber
\\
=&\ 
\E_{\tau\sim\P',\pi'}
\left[
\sum_{t=0}^\infty\gamma^t A^{\P,\pi'}(s_t,a_t)
\right]
+\underbrace{
\E_{\tau\sim\P,\pi'}\left[
\sum_{t=0}^\infty\gamma^t A^{\P,\pi'}(s_t,a_t)
\right]
-\E_{\tau\sim\P',\pi'}\left[
\sum_{t=0}^\infty\gamma^t A^{\P,\pi'}(s_t,a_t)
\right]
}_{\text{advantage discrepancy}}
-\frac{4\gamma\epsilon\delta_2^2}{(1-\gamma)^2}.
\label{eq:temp1}
\end{align}
}\noindent

We then analyze the above advantage discrepancy term. We have
{\small
\begin{align*}
&\E_{\tau\sim\P,\pi'}\left[
\sum_{t=0}^\infty\gamma^t A^{\P,\pi'}(s_t,a_t)
\right]
-\E_{\tau\sim\P',\pi'}\left[
\sum_{t=0}^\infty\gamma^t A^{\P,\pi'}(s_t,a_t)
\right]
\nonumber
\\ 
=&\ 
\sum_{t=0}^\infty\gamma^t\sum_{s_t}p^{\P,\pi'}(s_t)\sum_{a_t}\pi'(a_t|s_t)A^{\P,\pi'}(s_t,a_t)
-
\sum_{t=0}^\infty\gamma^t\sum_{s_t}p^{\P',\pi'}(s_t)\sum_{a_t}\pi'(a_t|s_t)A^{\P,\pi'}(s_t,a_t)
\nonumber
\\
=\ 
&\sum_{t=0}^\infty\gamma^t
\sum_{s_t}
\big(p^{\P,\pi'}(s_t)-p^{\P',\pi'}(s_t)\big)
\sum_{a_t}\pi'(a_t|s_t)A^{\P,\pi'}(s_t,a_t).
\end{align*}
}\noindent

Applying Lemma~\ref{proof:lemma1}, we can obtain
{\small
\begin{align}
&\left|
\E_{\tau\sim\P,\pi'}\left[
\sum_{t=0}^\infty\gamma^t A^{\P,\pi'}(s_t,a_t)
\right]
-\E_{\tau\sim\P',\pi'}\left[
\sum_{t=0}^\infty\gamma^t A^{\P,\pi'}(s_t,a_t)
\right]
\right|\leq
\frac{2\gamma\epsilon\delta_1}{(1-\gamma)^2}.
\label{eq:a_diff}
\end{align}
}\noindent

Combing Eq.~\eqref{eq:a_diff} and Eq.~\eqref{eq:temp1}, we have
{\small
\begin{align}
J(\P,\pi)-J(\P,\pi')\geq
\E_{\tau\sim\P',\pi'}\left[\sum_{t=0}^{\infty}\gamma^t A^{\P,\pi'}(s_t,a_t)\right]
-\frac{2\gamma\epsilon(\delta_1+2\delta_2^2)}{(1-\gamma)^2}.
\label{eq:mod_trpo}
\end{align}
}\noindent

Finally, combing Theorem~\ref{theo:2} and Eq.~\eqref{eq:mod_trpo}, we have
{\small
\begin{align}
&J(\P',\pi)-J(\P,\pi')
\nonumber
\\
=\ 
&J(\P',\pi)-J(\P,\pi)+J(\P,\pi)-J(\P,\pi')
\nonumber
\\
\geq\ 
&L_{\pi'}(\pi)+
\E_{\tau\sim\P',\pi'}
\left[
\sum_{t=0}^\infty\gamma^t A^{\P,\pi'}(s_t,a_t)
\right]-
\frac{2\gamma\epsilon(\delta_1+2\delta_2^2)}{(1-\gamma)^2}-
\frac{4\gamma r_{max}\delta_1}{(1-\gamma)^2}
\min\left(
\frac{\delta_2(\gamma^2+2)}{1-\gamma}, 1+\frac{\delta_2}{1-\gamma}
\right)
\nonumber
\\
=\ 
&\frac{1}{1-\gamma}\E_{s\sim d^{\P',\pi'} ,a,s'\sim\P',\pi'}\frac{\pi(a|s)}{\pi'(a|s)}
[r(s,a,s')+\gamma V^{\P,\pi'}(s')-Q^{\P,\pi'}(s,a)+A^{\P,\pi'}(s,a)]-C
\nonumber
\\
=\ 
&\frac{1}{1-\gamma}\E_{s\sim d^{\P',\pi'} ,a,s'\sim\P',\pi'}\frac{\pi(a|s)}{\pi'(a|s)}
[r(s,a,s')+\gamma V^{\P,\pi'}(s')-V^{\P,\pi'}(s)]-C,
\label{eq:27}
\end{align}
}\noindent
where 
$C=\frac{2\gamma\epsilon(\delta_1+2\delta_2^2)}{(1-\gamma)^2}
+
\frac{4\gamma r_{max}\delta_1}{(1-\gamma)^2}
\cdot\min\left(
\frac{\delta_2(\gamma^2+2)}{1-\gamma}, 1+\frac{\delta_2}{1-\gamma}
\right)$. 
Maximizing the lower bound in Eq.~\eqref{eq:27} is equivalent to
\begin{align}
\text{maximize}_{\pi}\ 
\E_{s\sim d^{\P',\pi'} ,a,s'\sim\P',\pi'}\frac{\pi(a|s)}{\pi'(a|s)}
\left[r(s,a,s')+\gamma V^{\P,\pi'}(s')\right],
\end{align}
by noting that $V^{\P,\pi'}(s)$ serves as a baseline and does not affect the policy gradient.
Now, we complete the proof.

\rightline{$\Box$}

\section{Proof of Theorem~\ref{theo:3}}
\label{sec:proof_theorem_rto}
\proof

Continuing from the relativity lemma, i.e., Eq.~(\ref{eq:diff}), we have
{\small
\begin{align*}
&J(\P',\pi)-J(\P,\pi)
\\
=\ &\E_{\tau\sim\P',\pi}\sum_{t=0}^{\infty}\gamma^t\ 
\left[
r(s_t,a_t,s_{t+1})+\gamma V^{\P,\pi}(s_{t+1})-Q^{\P,\pi}(s_t,a_t)
\right]
\\
=\ &\sum_{t=0}^{\infty}\gamma^t\E_{s_t,a_t\sim \P',\pi}\ 
\big[\E_{s_{t+1}\sim\P'(s_{t+1}|s_t,a_t)}
\left[
r(s_t,a_t,s_{t+1})+\gamma V^{\P,\pi}(s_{t+1})
\right]
\\
\ &\qquad\qquad\qquad\quad
-\E_{s_{t+1}\sim\P(s_{t+1}|s_t,a_t)}
\left[
r(s_t,a_t,s_{t+1})+\gamma V^{\P,\pi}(s_{t+1})
\right]
\big]
\\
=\ &\sum_{t=0}^{\infty}\gamma^t\E_{s_t,a_t\sim \P',\pi}\ 
\Bigg[
\sum_{s_{t+1}}\P'(s_{t+1}|s_t,a_t)\left(
r(s_t,a_t,s_{t+1})+\gamma V^{\P,\pi}(s_{t+1})\right)
\\
\ &\qquad\qquad\qquad\quad
-\sum_{s_{t+1}}\P(s_{t+1}|s_t,a_t)
\left(
r(s_t,a_t,s_{t+1})+\gamma V^{\P,\pi}(s_{t+1})
\right)
\Bigg].
\end{align*}
}\noindent
Now, considering a parameterized source dynamics function $\P_{\phi}$ and replacing $\P$ with $\P_{\phi}$, we have
{\small
\begin{align}
\Delta^{\P',\P_{\phi}}(\pi)
&=J(\P',\pi)-J(\P_{\phi},\pi)
\nonumber
\\
&=\sum_{t=0}^{\infty}\gamma^t\E_{s_t,a_t\sim \P',\pi}\ 
\Bigg[
\sum_{s_{t+1}}\P'(s_{t+1}|s_t,a_t)\left(
r(s_t,a_t,s_{t+1})+\gamma V^{\P_{\phi},\pi}(s_{t+1})\right)
\nonumber
\\
&\qquad\qquad\qquad\qquad\quad
-\sum_{s_{t+1}}\P_{\phi}(s_{t+1}|s_t,a_t)
\left(
r(s_t,a_t,s_{t+1})+\gamma V^{\P_{\phi},\pi}(s_{t+1})
\right)
\Bigg].
\label{eq:proof_rto1}
\end{align}
}\noindent

Let
{\small
\begin{align*}
L_{\phi'}(\phi)=
&\sum_{t=0}^{\infty}\gamma^t\E_{s_t,a_t\sim \P',\pi}\ 
\Bigg[\sum_{s_{t+1}}\P'(s_{t+1}|s_t, a_t)\left(
r(s_t,a_t,s_{t+1})+\gamma V^{\P_{\phi'},\pi}(s_{t+1})\right)
\\
&\qquad\qquad\qquad\quad
-\sum_{s_{t+1}}\P_{\phi}(s_{t+1}|s_t,a_t)
\left(
r(s_t,a_t,s_{t+1})+\gamma V^{\P_{\phi'},\pi}(s_{t+1})
\right)
\Bigg]
\end{align*}
be an approximate function of $\Delta^{\P',\P_{\phi}}(\pi)$ by replacing $V^{\P_{\phi},\pi}$ in Eq.~\eqref{eq:proof_rto1} with $V^{\P_{\phi'},\pi}$, i.e., importing another dynamics parameter $\phi'$ to evaluate the value. Then,
\begin{align*}
&\Delta^{\P',\P_{\phi}}(\pi)-L_{\phi'}(\phi)=
\\
&\sum_{t=0}^{\infty}\gamma^{t+1}\E_{s_t,a_t\sim \P',\pi}\left[
\sum_{s_{t+1}}
\left(
\P'(s_{t+1}|s_t,a_t)-\P_{\phi}(s_{t+1}|s_t,a_t)
\right)
\left(
V^{\P_{\phi},\pi}(s_{t+1})-V^{\P_{\phi'},\pi}(s_{t+1})
\right)
\right].
\end{align*}
}\noindent

Applying Lemma~\ref{proof:lemma2}, we obtain the bound
{\small
\begin{align*}
&|\Delta^{\P',\P_{\phi}}(\pi)-L_{\phi'}(\phi)|
\\
\leq\ &
\sum_{t=0}^{\infty}\gamma^{t+1}\E_{s_t,a_t\sim \P',\pi}\left[
\sum_{s_{t+1}}
\left|
\P'(s_{t+1}|s_t,a_t)-\P_{\phi}(s_{t+1}|s_t,a_t)
\right|\cdot\frac{2r_{max}}{1-\gamma}
\min\left(\delta_1\left(
t\gamma+\frac{1}{1-\gamma}
\right), 1
\right)
\right]
\\
\leq\ &
\frac{4\gamma\delta_1r_{max}}{(1-\gamma)^2}
\min\left(\frac{\delta_1(\gamma^2+1)}{1-\gamma}, 1\right).
\end{align*}
}\noindent
Alternatively, we can also use the discrepancy between the dynamics parameters $\phi$ and $\phi'$ to bound the above equation instead of using Lemma~\ref{proof:lemma2}. When taking small step sizes in updating $\phi$ to $\phi'$, we can obtain sufficiently tight bound, because the above bound only depends on $\delta_1$, i.e., the differences between $\phi$ and $\phi'$ here.

\rightline{$\Box$}

\section{The RPO and RTO Algorithms}
Due to space limitations, we are not able to put the RPO and RTO algorithms in the main text. These two algorithms can be tailored from the RPTO algorithm in Algorithm~\ref{alg:rpto}, and we put them in Algorithm~\ref{alg:rpo} and Algorithm~\ref{alg:rto} respectively.

\begin{algorithm}[h]
\small
   \caption{Relative Policy Optimization (RPO)}
   \label{alg:rpo}
\begin{algorithmic}
   \STATE {\bfseries Input:} the source and target environments $\mathcal{E}^{source}$ and $\mathcal{E}^{target}$, and their dynamics $\P^{source}$ and $\P^{target}$; a well-trained policy $\pi_{\theta_0}$ in $\mathcal{E}^{source}$\;
   \STATE \textbf{1.} Create two empty replay buffers $\mathcal{D}_{source}$ and $\mathcal{D}_{target}$\;
   \STATE \textbf{2.} Initialize $\theta=\theta_0$\;
   \REPEAT
   \STATE \textbf{3.} Using $\pi_{\theta}$ to interact with $\mathcal{E}_{\phi}^{source}$ and push the generated trajectories into $\mathcal{D}_{source}$\;
   \STATE \textbf{4.} Using $\pi_{\theta}$ to interact with $\mathcal{E}^{target}$ and push the generated trajectories into $\mathcal{D}_{target}$\;
   \STATE \textbf{5.} Sample a mini-batch $\{(s,a,s')\}_{source}\sim\mathcal{D}_{source}$, and update $Q^{\P_{\phi}^{source},\pi_{\theta}}$ by minimizing the soft Bellman residual\;
   \STATE \textbf{6.} Sample a mini-batch $\{(s,a,s')\}_{target}\sim\mathcal{D}_{target}$, and apply the relative policy gradient in RPO to update $\pi_{\theta}$\;
   \UNTIL{Some convergence criteria is satisfied}
\end{algorithmic}
\end{algorithm}

\begin{algorithm}[h]
\small
   \caption{Relative Transition Optimization (RTO)}
   \label{alg:rto}
\begin{algorithmic}
   \STATE {\bfseries Input:} the source and target environments $\mathcal{E}^{source}$ and $\mathcal{E}^{target}$, and their dynamics $\P_{\phi}^{source}$ and $\P^{target}$, where the source dynamics $\P_{\phi}^{source}$ is parameterized by $\phi$; an arbitrary policy $\pi_{\theta}$\;
   \STATE \textbf{1.} Create two empty replay buffers $\mathcal{D}_{source}$ and $\mathcal{D}_{target}$\;
   \STATE \textbf{2.} Initialize $\phi=\phi_0$\;
   \REPEAT
   \STATE \textbf{3.} Using $\pi_{\theta}$ to interact with $\mathcal{E}_{\phi}^{source}$ and push the generated trajectories into $\mathcal{D}_{source}$\;
   \STATE \textbf{4.} Using $\pi_{\theta}$ to interact with $\mathcal{E}^{target}$ and push the generated trajectories into $\mathcal{D}_{target}$\;
   \STATE \textbf{5.} Sample a mini-batch $\{(s,a,s')\}_{source}\sim\mathcal{D}_{source}$, and update $Q^{\P_{\phi}^{source},\pi_{\theta}}$ by minimizing the soft Bellman residual\;
   \STATE \textbf{6.} Sample a mini-batch $\{(s,a,s')\}_{target}\sim\mathcal{D}_{target}$, and update $\P_{\phi}^{source}$ according to RTO or SL\;
   \UNTIL{Some convergence criteria is satisfied}
\end{algorithmic}
\end{algorithm}

\section{Other Experimental Details}
\label{sec:G}
For the experiments on MuJoCo tasks, we use 2 fully connected layers with 256 units to build policy and value neural networks respectively. The policy network outputs the mean and covariance of Gaussian distribution. We build the neural dynamics model following MBPO~\cite{janner2019trust}, and use an ensemble network to learn the dynamics.
We set the ensemble size to 7 and each ensemble network has 4 fully connected layers with 400 units.
Each head of the dynamics model is a probabilistic neural network which outputs the Gaussian distribution with diagonal covariance: $P_{\phi}^i(s_{t+1}, r_t|, s_t, a_t) = \mathcal{N}\big(\mu_{\phi}^i(s_t, a_t), \sum_{\phi}^i(s_t,a_t)\big)$. We set the model horizon to 1 and the replay ratio of dynamic to 1 for all environments.
We select the replay ratio of policy $\mathrm{RR}_p$ and RTO minimum weight $\epsilon$ for each environment by sweeping hyper-parameter. Detailed settings are shown in Table~\ref{tb:setting_mujoco}. 
As mentioned in Section~\ref{sec:rpo}, we need to use the standard SAC loss to keep the policy performing well in the source environment. In our implementation, we alternately sample data from $\mathcal{D}_{source}$ and $\mathcal{D}_{target}$ to update the policy. We set this alternate frequency $f$ to 5 for all environments.
In other words, we sample 1 batch data from $\mathcal{D}_{source}$ after every 4 batch data from $\mathcal{D}_{target}$.
All other settings are the same as the original MBPO~\cite{janner2019trust}.

\begin{table}[h]
\caption{Detailed algorithm setting for MuJoCo tasks.}
\label{tb:setting_mujoco}
\centering
\begin{tabular}{l|ccccc}
\hline
                 & Ant & HalfCheetah & Hopper & Walker2d & Swimmer \\
\hline
 $\mathrm{RR}_p$ & 5   & 1   & 2   &  1    & 20\\
 $\epsilon$      & 1.0 & 0.5 & 1.0 &  0.5  & 0.5     \\
\hline
\end{tabular}
\end{table}

For environment settings, we first modify the Ant and Swimmer in OpenAI as our source environment.
For the Ant, we clip the observation space from 111 to 27 following MBPO~\cite{janner2019trust}, 
which makes it easier to learn the dynamics model.
For the Swimmer, we add 4 more joints to the default environment, which creates additional 2 actions.
For the design of the target environment, we randomly add $\pm20\%$ noise to the length of joint and friction.
For every environment, we create 4 different target environments, and we repeat 4 times for target environments.
Therefore, we repeat 16 times for each task to plot the results in Figure~\ref{fig:mujoco}.

All experiments were ran with one NVIDIA TESLA M40 GPU and tens of CPU cores.

\end{document}